\documentclass[journal]{IEEEtran}

\ifCLASSINFOpdf
\else
\fi
\usepackage{times}
\usepackage{epsfig}
\usepackage{graphicx}
\usepackage{amsmath}
\usepackage{amssymb}
\usepackage{multirow}
\usepackage{gensymb}
\usepackage{bm,color}
\usepackage{caption}
\usepackage{graphicx}
\usepackage{float}
\usepackage{subfigure}
\graphicspath{{./figs/}}
\renewcommand{\mathbf}{\boldsymbol}
\newcommand{\tx}{\bm{x}}

\newcommand{\tb}{\bm{b}}
\newcommand{\te}{\bm{e}}
\newcommand{\tv}{\bm{v}}

\newcommand{\ty}{\bm{y}}
\newcommand{\tA}{\bm{A}}
\newcommand{\tW}{\bm{W}}
\newcommand{\tw}{\bm{w}}
\newcommand{\tX}{\bm{X}}
\newcommand{\tS}{\bm{S}}
\newcommand{\tn}{\bm{n}}
\newcommand{\tbh}{\bm{h}}
\newcommand{\tI}{\bm{I}}

\newcommand{\tOmega}{\mathbf{\Omega}}

\DeclareMathOperator*{\argmin}{argmin}

\usepackage{adjustbox}

\newcommand{\widthscalefive}{0.165}
\newcommand{\widthscaleeight}{0.115}
\begin{document}

\title{Accurate and Lightweight Image Super-Resolution with Model-Guided Deep Unfolding Network}

\author{Qian Ning,~\IEEEmembership{Student Member,~IEEE,}
        Weisheng Dong,~\IEEEmembership{Member,~IEEE,}
        Guangming Shi,~\IEEEmembership{Senior Member,~IEEE,}
	   Leida~Li,~\IEEEmembership{Member,~IEEE}
        and~Xin Li,~\IEEEmembership{Fellow,~IEEE}
\thanks{Qian Ning, Weisheng Dong, Guangming Shi and Leida Li are with School of Artificial Intelligence, Xidian University, Xi'an, 710071, China.}
\thanks{Xin Li is with the Lane Department of Computer Science and Electrical Engineering, West Virginia University, Morgantown WV 26506-6109.
}
\thanks{This work was supported in part by the National Key R\&D Program of China under Grant 2018AAA0101400 and the Natural Science Foundation of China under Grant 61991451, Grant 61632019, Grant 61621005, and Grant 61836008. Xin Li's work is partially supported by the DoJ/NIJ under grant NIJ 2018-75-CX-0032, NSF under grant OAC-1839909 and the WV Higher Education Policy Commission Grant (HEPC.dsr.18.5).}}
%
%

\maketitle

\begin{abstract}
Deep neural networks (DNNs) based methods have achieved great success in single image super-resolution (SISR). However, existing state-of-the-art SISR techniques are designed like black boxes lacking transparency and interpretability. Moreover, the improvement in visual quality is often at the price of increased model complexity due to black-box design. In this paper, we present and advocate an explainable approach toward SISR named model-guided deep unfolding network (MoG-DUN). Targeting at breaking the coherence barrier, we opt to work with a well-established image prior named nonlocal auto-regressive model and use it to guide our DNN design. By integrating deep denoising and nonlocal regularization as trainable modules within a deep learning framework, we can unfold the iterative process of model-based SISR into a multi-stage concatenation of building blocks with three interconnected modules (denoising, nonlocal-AR, and reconstruction). The design of all three modules leverages the latest advances including dense/skip connections as well as fast nonlocal implementation. In addition to explainability, MoG-DUN is accurate (producing fewer aliasing artifacts), computationally efficient (with reduced model parameters), and versatile (capable of handling multiple degradations). The superiority of the proposed MoG-DUN method to existing state-of-the-art image SR methods including RCAN, SRMDNF, and SRFBN is substantiated by extensive experiments on several popular datasets and various degradation scenarios.
\end{abstract}

\IEEEpeerreviewmaketitle
\section{Introduction}
\label{sec:1}

The field of deep learning for single image super-resolution (SISR) has advanced rapidly in recent years. Super-resolution by convolutional neural network (SRCNN) \cite{dong:ECCV:2014learning} represented one of the pioneering works in this field.
Since then, many follow-up works have been developed including Super-resolution via Generative Adversarial Network (SRGAN) \cite{ledig2017photo}, SR via very deep convolutional networks (VDSR) \cite{kim2016accurate}, Trainable Nonlinear Reaction Diffusion Network(TNRD)~\cite{chen:TPAMI:2017TNRD}, Deeply-recursive convolutional network (DRCN) \cite{kim:CVPR:2016DRCN}, Enhanced Deep Residual Networks (EDSR) \cite{lim:CVPR:2017:EDSR}, Laplacian Pyramid
Super-Resolution Network  (LapSRN) \cite{lai2017deep}, and Deep Back-Projection Networks (DBPN) \cite{haris2018deep}. Most recently, SISR has benefited from the advances in novel design of network architectures such as densely-connected networks (e.g., RDN \cite{zhang:CVPR:2018RDN}), attention mechanism (e.g., RCAN \cite{zhang:ECCV:2018RCAN} ans SAN \cite{dai2019second}), multiple degradations (e.g., SRMDNF \cite{zhang:CVPR:2018SRMDNF}), feedback connections (e.g., SRFBN \cite{li:CVPR:2019feedback} and feature aggregation \cite{liu2020residual}.


Despite the rapid progress, one of the long-standing open issues is the lack of interpretability. Most existing networks for SISR are designed based on the black-box principle - i.e., little is known about their internal workings regardless of the desirable input-output mapping results. The difficulty with understanding the internal mechanism of deep learning-based SISR has become even more striking when the network gets deeper and more sophisticated (e.g., due to attention and feedback). For instance, the total number of parameters of EDSR \cite{lim:CVPR:2017:EDSR} has reached over 40M, which makes it a less feasible for practical applications. By contrast, it is more desirable to seek alternative glass-box design (a.k.a. clear-box) where the inner components are readily available for inspection. A hidden benefit of such a transparent approach is that it might lead to more efficient design because any potential redundancy (in terms of model parameters) can be cautiously avoided. Can we solve SISR under an emerging framework of interpretable machine learning \cite{molnar2020interpretable}? Does a transparent design lead to computationally more efficient solution to SISR facilitating practical applications (e.g., lightweight architecture \cite{ahn2018fast})?

We provide affirmative answers to the above questions in this paper. An important new insight brought to the field of SISR by this work is {\it model-guided} (MoG) design for deep neural networks. The basic idea behind MoG design is to seek a mathematically equivalent implementation of existing model-based solution by deep neural networks. Similar ideas have scattered in the literature of so-called deep unfolding networks (e.g., \cite{wisdom2017building,bertocchi2019deep,hershey2014deep}). It has been well-established that classic model-based tools including sparse coding, Markov Random Field, belief propagation, and non-negative matrix factorization can be unfolded to a network implementation \cite{hershey2014deep,wisdom2017building,dong:TPAMI:2018DPDNN,zhang:CVPR:2020deepUSRNet}. Note that in the field of SISR, there are plenty of model-based methods \cite{buades:CVPR:2005nonlocal,yang:TIP:2010:sparse,dong:TIP:2013nonlocally,kim:TPAMI:2010sparse}, which could provide a rich source of inspiration for MoG design.  Most recent work USRNet \cite{zhang:CVPR:2020deepUSRNet}  serves as an exemplar of such MoG design for  handling classical degradation of SISR via a single unified model.  Based on our previous \cite{dong2013sparse} and recent \cite{dong:TPAMI:2018DPDNN} works, we propose a new MoG Deep Unfolding Network (MoG-DUN) for SISR that is not only explainable and efficient but also accurate and versatile.
\begin{figure}[htbh]
\centering
\includegraphics[width=0.95\linewidth]{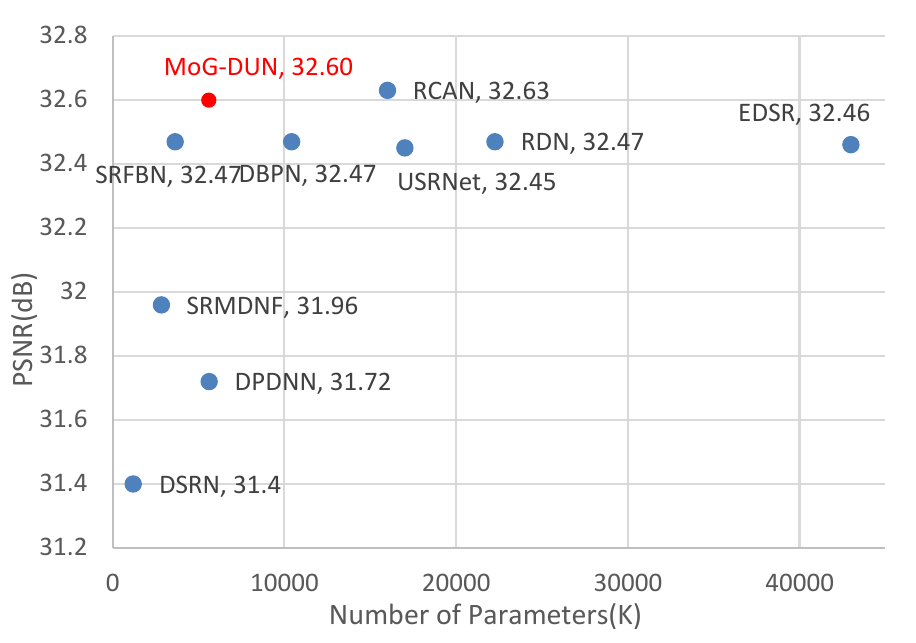}
\caption{This work advances the state-of-the-art in SISR by striking an improved tradeoff between the cost (as measured by the number of model parameters) and performance (reflected by the PSNR metric on Set5 with $\times4$ bicubic down-sampling). Our methods are highlighted by the red color. }
  \label{fig:parameters}
\end{figure}

We will show that most model-based image priors or regularization functions, including both convex and nonconvex formulation, can be leveraged as the guidance for the design of deep neural networks. In fact, the class of nonconvex regularization models do not pose additional difficulty to learning-based approaches despite their intractability from an analytical perspective. This is because that the training of deep neural networks (DNN) has a natural and intrinsic connection with the optimization of cost functions by numerical methods (e.g., gradient descent methods \cite{le2011optimization}). Aiming at breaking the coherence barrier \cite{adcock2013breaking} (a long-standing open problem in SISR), we have chosen a reference model, nonlocal autoregressive model (NARM) with improved incoherence properties \cite{dong2013sparse}, to showcase the process of Model-Guided (MoG) design. We will unfold this model with a nonconvex cost function into network implementations consisting of multistage U-net modules \cite{dong:TPAMI:2018DPDNN}. This work further extends our previous work \cite{dong:TPAMI:2018DPDNN} by incorporating a nonlocal module (for AR modeling computation) and a reconstruction module (for AR modeling correction). Moreover, long connections are introduced across different stages to copy the hidden states from previous stages to the current, which facilitates the information flow. Even though MoG-DUN has achieved outstanding performance (in terms of both subjective and objective qualities), its model complexity has been kept much lower than that of DBPN \cite{haris2018deep}, RDN \cite{zhang:CVPR:2018RDN}, and RCAN \cite{zhang:ECCV:2018RCAN} and only slightly higher than that of SRMDNF \cite{zhang:CVPR:2018SRMDNF} and SRFBN \cite{li:CVPR:2019feedback}, as shown in Fig. \ref{fig:parameters}. Meantime, MoG-DUN is explainable and versatile thanks for its transparent design, which has the potential of leveraging to other image restoration applications.

The key contributions of this paper are summarized below.

\begin{itemize}
    \item We present a model-guided explainable approach toward SISR. Our approach is capable of leveraging existing NARM-based SISR into a network implementation in a transparent manner. Thanks to the improved incoherence property of NARM, the corresponding MoG-DUN has better capability of alleviating long-standing problem in SISR such as aliasing artifacts. The framework of MoG design is applicable to all existing models including nonconvex and nonlocal regularization. 

\item  We demonstrate how to unfold an existing NARM \cite{dong2013sparse} model into the corresponding network implementation. The unfolding result consists of the multi-stage concatenation of Unet-like deep denoising module along with a nonlocal-AR module and a reconstruction module. Both long and short skip connections are introduced within and across different stages to facilitate the information flow. The model complexity of our MoG-DUN is shown to be comparable to that of \cite{dong:TPAMI:2018DPDNN}.

\item We report extensive experimental results for the developed MoG-DUN, which justify its achieving an improved trade-off between the modeling complexity and the SR reconstruction performance. Our MoG-DUN has achieved better performance than existing state-of-the-art SISR such as RDN \cite{zhang:CVPR:2018RDN}) and RCAN \cite{zhang:ECCV:2018RCAN} with a lower cost as measured by the number of model parameters. In particular, visual quality improvement in terms of sharper edges and more faithful reconstruction of texture patterns is mostly striking.
\end{itemize}


\section{Model-based Image Interpolation and Restoration}
\label{sec:2}

We first briefly review previous works on model-based image interpolation (e.g.,  NARM \cite{dong2013sparse}) and image restoration (e.g., DPDNN \cite{dong:TPAMI:2018DPDNN}), which sets up the stage for model-guided network design. The NARM model has achieved state-of-the-art performance in model-based image interpolation (a special case of SISR involving down-sampling only and without any anti-aliasing low-pass filter). Denoising Prior driven DNN (DPDNN) \cite{dong:TPAMI:2018DPDNN} represents the most recent MoG design of deep unfolding network whose variation has been adopted as the baseline method in our research.

\begin{figure}[htbh]
\begin{center}
\includegraphics[height=1.8in]{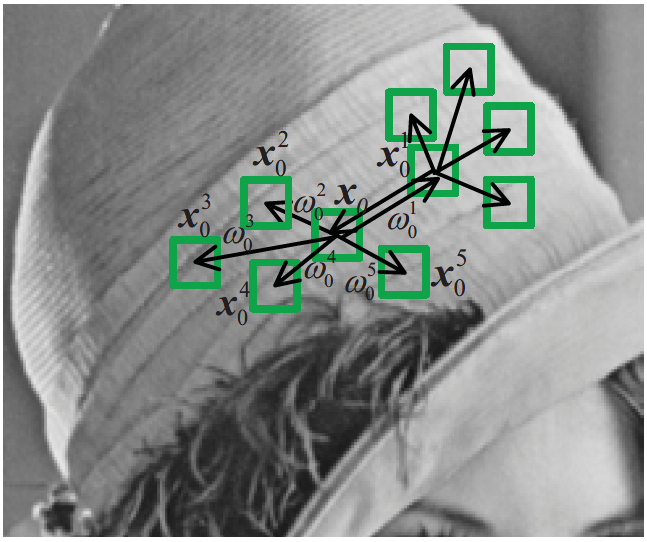}
\end{center}
\caption{The illustration of NARM \cite{dong2013sparse} - a nonlocal extension of classic auto-regressive (AR) model for image signals.}
\label{fig:NARM}
\end{figure}

\subsection{Nonlocal Auto-regressive Model (NARM)}

\begin{figure*}[htbh]
\newlength\structure
\setlength{\structure}{-5mm}
\centering
\begin{tabular}{c}
		\begin{adjustbox}{valign=t}
		\tiny
			\begin{tabular}{c}
                \includegraphics[width=1.0\linewidth]{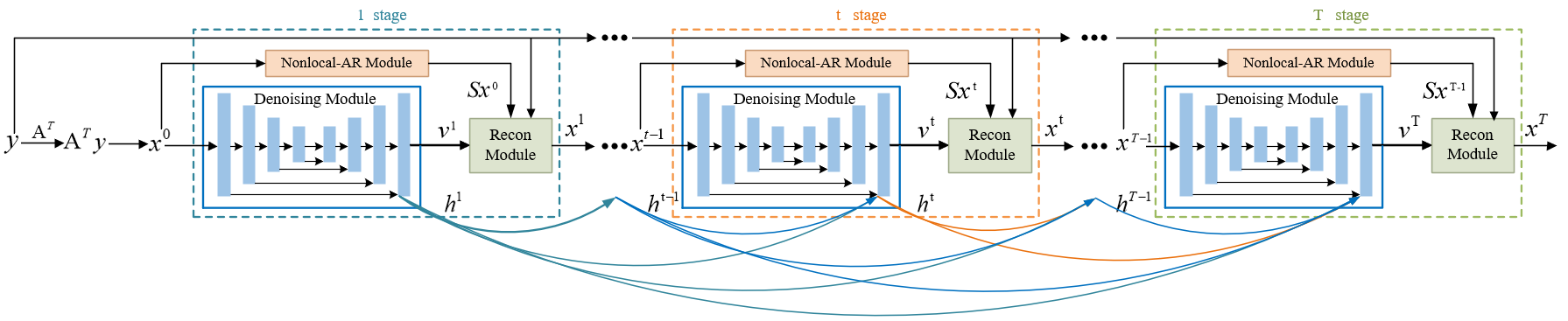}
                \\
                (a) The overview of our proposed MoG-DUN network for SISR.
            \end{tabular}
		\end{adjustbox}
        \\
        \begin{adjustbox}{valign=t}
		\tiny
			\begin{tabular}{cccccc}

            \includegraphics[height=1.5in]{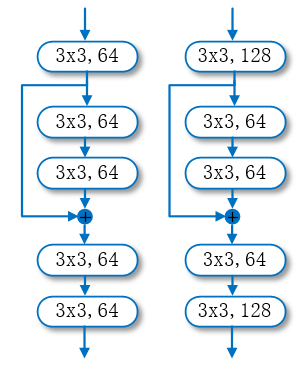} \hspace{\structure} &
            \includegraphics[height=1.5in]{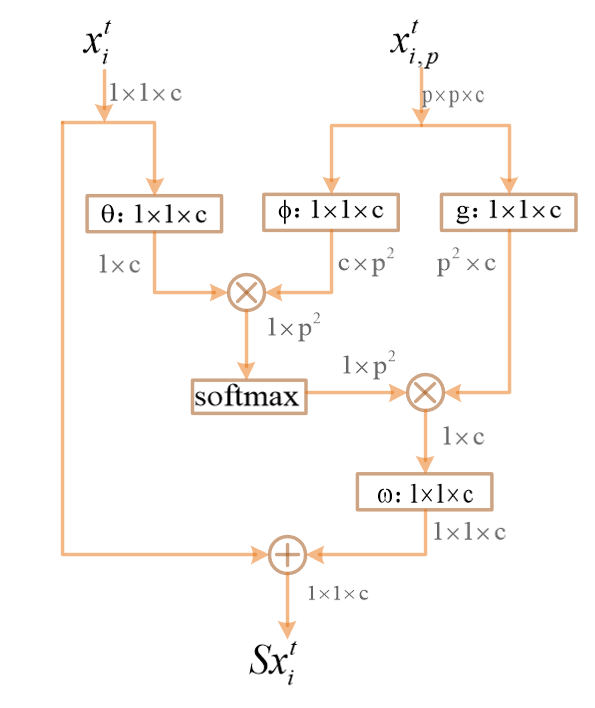} \hspace{\structure} &
            \includegraphics[height=1.5in]{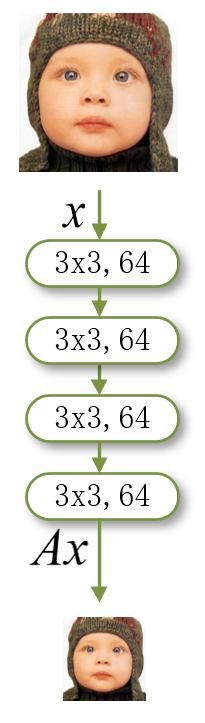} \hspace{\structure} &
            \includegraphics[height=1.5in]{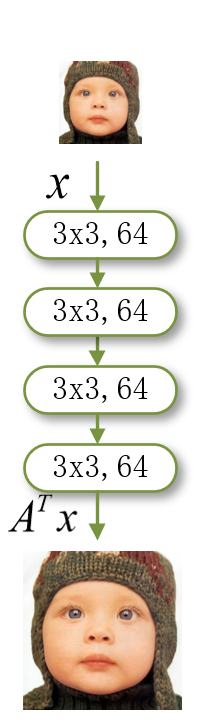} \hspace{\structure} &
            \includegraphics[height=1.5in]{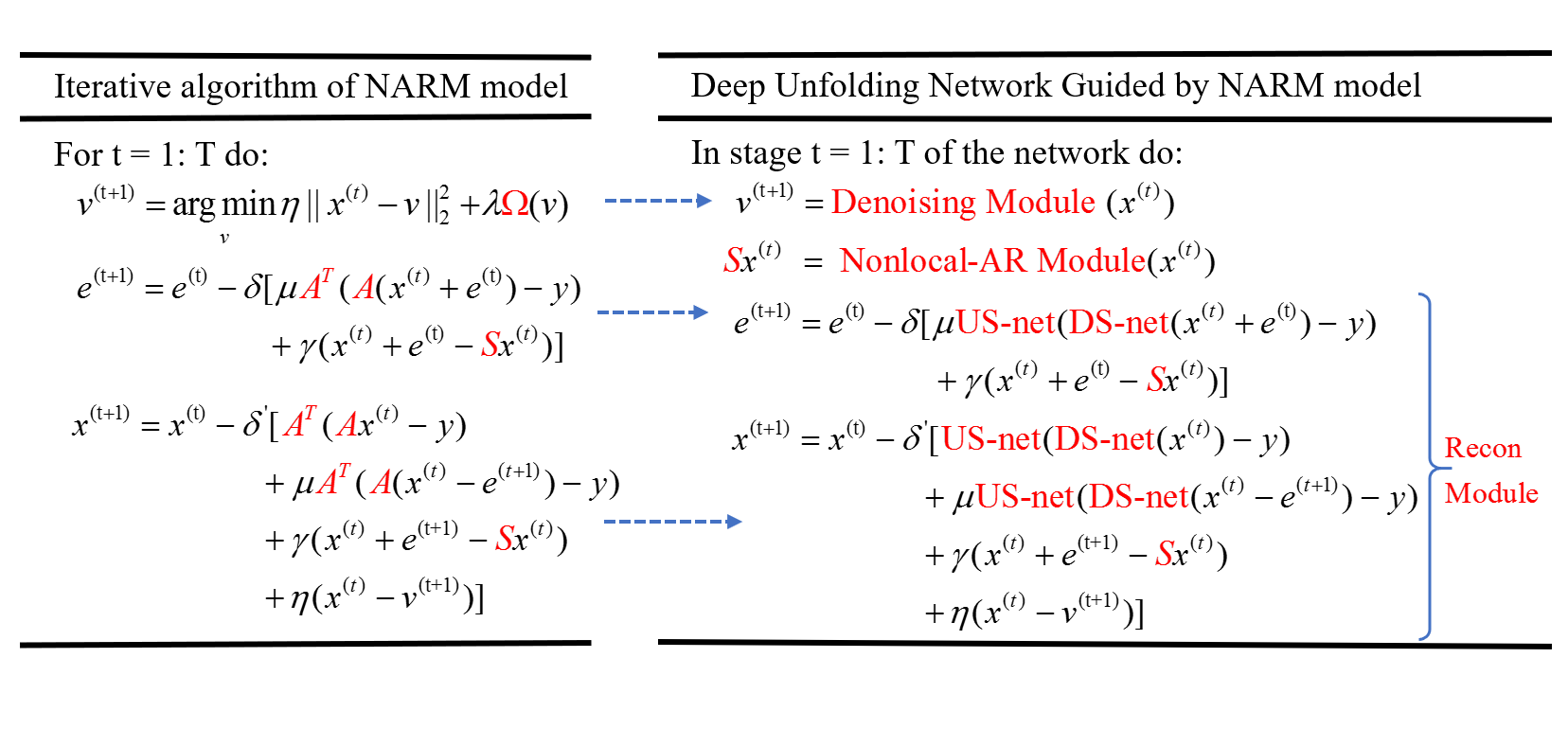} \hspace{\structure} &
            \\
             (b) \emph{Deep denoising module} \hspace{\structure} &
			 (c) \emph{Fast Nonlocal-AR module} \hspace{\structure} &
			 (d) \emph{DS-net} \hspace{\structure} &
             (e) \emph{US-net} \hspace{\structure} &
			 (f) \emph{The illustration of relationship between the Iterative algorithm of NARM model and MoG-DUN Network }\hspace{\structure} &
         \end{tabular}
		\end{adjustbox}
\end{tabular}
  \caption{The architecture of our proposed MoG-DUN network for SISR. (a) The overall architecture of the proposed network($\tbh^{t-1},\tbh^{t},\tbh^{T-1}$ denote the hidden states of corresponding stages), (b)The inner structure of the Encoding and Decoding block of the U-net \textbf{Denoising Module}, (c) The architecture of the \textbf{Nonlocal-AR Module}, (d) The inner structure of \textbf{Down-Sampling-network(DS-net)}, (e) The inner structure of \textbf{Up-Sampling-network(US-net)}, (f) The illustration of corresponding relationship between the Iterative algorithm of NARM model and MoG-DUN Network. }
  \label{fig:framework}
\end{figure*}

The basic idea behind nonlocal auto-regressive modeling (refer to Fig. \ref{fig:NARM}) is to extend the traditional auto-regressive (AR) models by redefining the neighborhood. For a given patch $\tx_i$, NARM seeks its sparse linear decomposition over a set of nonlocal (instead of local) neighborhood. Following the notation in \cite{dong2013sparse}, we have
\begin{equation}
\tx_i \approx \sum_j \omega_i^j \tx_i^j \label{NARM_1}
\end{equation}
where $\tx_i^j$ denotes the $j$-th similar patch found in the nonlocal neighborhood. A natural way of extending the classic AR modeling is to formulate the following regularized Least-Square problem:
\begin{equation}
\tw_i = \argmin_{\tw_i} ||\tx_i-\tX \tw_i||_2^2+\gamma ||\tw_i||_2^2 \label{NARM_2}
\end{equation}
where $\tX=[\tx_i^1,\tx_i^2,...,\tx_i^J]$, $\tw_i=[\omega_i^1,\omega_i^2,...,\omega_i^J]^T$, and $\gamma$ is the regularization parameter. The closed-form solution to Eq. \eqref{NARM_2} is given by
\begin{equation}
\tw_i = (\tX^T\tX + \gamma \tI)^{-1} (\tX^T \tx_i)\label{NARM_3}
\end{equation}
where $\tI$ is an identity matrix with the same size as $\tX^T\tx$. Based on newly determined AR coefficients $\tw_i$, we can represent
the nonlocal autoregressive model (NARM) of image $x$ by
\begin{equation}
\tx = \tS\tx+\te_x \label{NARM_eq}
\end{equation}
where $\te_x$ is the modeling error, and the NARM matrix $\tS$ is
\begin{equation}
\tS_{i,j} = \left \{ \begin{array}{ll}
\omega_i^j,& \textrm{if $\tx_i^j$ is a nonlocal neighbor of $\tx_i$} \\
0,   &\textrm{otherwise} \label{NARM_S}
\end{array} \right.
\end{equation}
The NARM matrix $\tS$ can be embedded into a standard image degradation model by modifying Eq. \eqref{degration} into
\begin{equation}
\ty = \tilde{\tA} \tx + \tn \label{degration2}
\end{equation}
where $\tilde{\tA} =\tA \tS$ is the new degradation operator. It has been shown in \cite{dong2013sparse} that such nonlocal extension is beneficial to improving the incoherence between sampling matrix and sparse dictionaries under the framework of model-based image restoration. In this work, we will show how to unfold this NARM into a DNN-based implementation.

\subsection{Model-based Image Restoration}
The objective of model-based image restoration (IR) is to estimate an unknown image $\tx$ from its degraded observation $\ty$. The degradation process can be formulated by:
\begin{equation}
\ty = \tA \tx + \tn \label{degration}
\end{equation}
where $\tA$ denotes the degradation operators (e.g., blurring kernels, down-sampling operations) and $\tn$ denotes the additive noise. Accordingly, model-based IR can be formulated into the following optimization problem:
\begin{equation}
\tx = \argmin_{\tx}||\ty-\tA \tx||_2^2 + \lambda\tOmega(\tx) \label{model-based}
\end{equation}
where $\lambda$ is the Lagrangian multiplier and $\tOmega(\tx)$ the regularization  function. The choice of various regularization  functions $\tOmega(\tx)$ reflects
different ways of incorporating a priori knowledge about the unknown HR image $\tx$.

To solve model-based SR problem in Eq. (\ref{model-based}), half-quadratic splitting method \cite{he2013half} converts an equally-constrained optimization problem into an equivalent non-constrained optimization problem, which can be written as
\begin{equation}
(\tx,\tv) = \argmin_{\tx,\tv}||\ty-\tA \tx||_2^2 +\eta||\tx-\tv||_2^2
+ \lambda\tOmega(\tv) \label{model-based_problem}
\end{equation}
where $\tv$ is an auxiliary splitting variable. It follows that Eq. \eqref{model-based_problem} boils down to alternatively solving two sub-problems associated with the fidelity and regularization terms respectively,
\begin{subequations}
\begin{equation}\tx^{(t+1)} = \argmin_{\tx}||\ty-\tA \tx||_2^2 +\eta||\tx-\tv^{(t)}||_2^2 \label{sub_problem_1} \end{equation}
\begin{equation}\tv^{(t+1)} = \argmin_{\tv}\eta||\tx^{(t+1)}-\tv||_2^2+\lambda\tOmega(\tv) \label{sub_problem_2} \end{equation}
\end{subequations}

The main idea behind deep unfolding network is that conventional iterative soft-thresholding algorithm (ISTA) in sparse coding can be implemented equivalently by a stack of recurrent neural networks \cite{wisdom2017building}. Such correspondence has inspired a class of convolutional neural network (CNN)-based image denoising techniques \cite{zhang2017beyond}. CNN-based denoising prior was later extended into other model-based image restoration problems in Image Restoration via CNN (IRCNN) \cite{zhang:CVPR:2017learning}. In IRCNN, a CNN module is adopted to solve Eq.(\ref{sub_problem_2}) while Eq.(\ref{sub_problem_1}) is solved in close-form by
\begin{equation}
\tx^{(t+1)}=\tW^{-1}\tb,
\end{equation}
where $\tW$ represents the matrix related to the degradation matrix $\tA$ and b is referring to  $\eta\tv^{(t)}+\tA^T$. Note that it is often time-consuming to calculate an inverse matrix. Based on such observation, a computationally more efficient unfolding strategy was developed in DPDNN \cite{dong:TPAMI:2018DPDNN}. DPDNN addressed the problem of matrix inverse by solving Eq.(\ref{sub_problem_1}) in a different way - i.e., they propose to compute $\tx^{(t+1)}$ with a single step of gradient descent by
\begin{equation}
\begin{split}
\tx^{(t+1)} &= \tx^t-\delta[\tA^T(\tA\tx^{(t)}-\ty) + \eta(\tx^{(t)}-\tv^{(t)})]\\
&=\bar{\tA}\tx^{(t)}+\delta \tA^T\ty+\delta\eta \tv^{(t)}
\end{split}
\label{DPDNN_solvtion}
\end{equation}
where $\bar{\tA}=[(1-\delta\eta)\tI-\delta \tA^T\tA]$ can be pre-computed (so the updating of $\tx^{(t)}$ can be done efficiently). As shown in \cite{dong:TPAMI:2018DPDNN}, we only need to obtain an approximated solution to the $\tx$-subproblem, which can be done in a computationally efficient manner. Although DPDNN \cite{dong:TPAMI:2018DPDNN} has found several image restoration applications including SISR, its network architecture remains primitive (e.g., lacking dense and skip connections) and there is still room for further optimization.

\section{Model-guided Deep Unfolding Network}
\label{sec:3}

In this paper, we take another step forward by showing a generalized unfolding strategy applicable to almost any model-based image restoration including both nonlocal and nonconvex regularization. It is well known that the class of nonconvex optimization problems do not admit computationally efficient solutions, which calls for convex relaxation or approximation \cite{saab2008stable}. However, we note that nonconvex cost functions have become almost the default option in the field of machine learning \cite{jain2017non} especially deep learning \cite{goodfellow2016deep}. This is likely due to the fact that the training of deep neural networks (DNN) has a natural and intrinsic connection with the optimization of cost functions by numerical methods such as gradient descent methods \cite{le2011optimization} (e.g., the popular Adam optimization represents a stochastic gradient descent method). Here, we have chosen a previous reference model (NARM originally designed for image interpolation \cite{dong2013sparse}) to showcase the process of Model-Guided (MoG) design. Thanks to the improved coherence property of NARM, unfolding this model has the potential of alleviating some long-standing open problems in SISR such as the suppression of aliasing artifacts. 

\begin{figure*}[htbp]
	\newlength\fsdttwofigBD
	\setlength{\fsdttwofigBD}{-1.5mm}
	\centering
	\begin{tabular}{cc}
		\begin{adjustbox}{valign=t}
		\tiny
			\begin{tabular}{c}
				\includegraphics[width=0.233\textwidth]{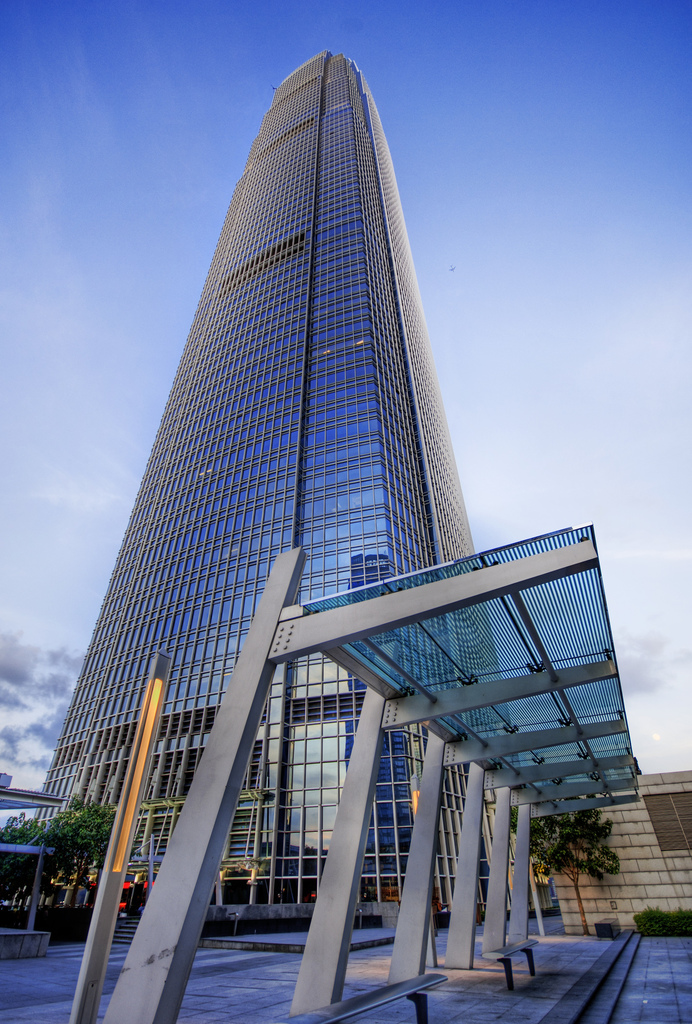}
				\\
				 \emph{Img\_046}~ from Urban100 $\times4$
				
				
			\end{tabular}
		\end{adjustbox}
		\hspace{-2.3mm}
		\begin{adjustbox}{valign=t}
		\tiny
			\begin{tabular}{cccccc}
				\includegraphics[width=\widthscalefive \textwidth]{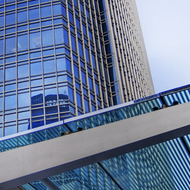} \hspace{\fsdttwofigBD} &
				\includegraphics[width=\widthscalefive \textwidth]{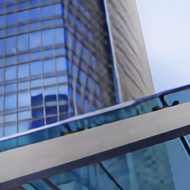} \hspace{\fsdttwofigBD} &
				\includegraphics[width=\widthscalefive \textwidth]{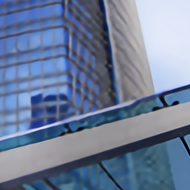} \hspace{\fsdttwofigBD} &
				\includegraphics[width=\widthscalefive \textwidth]{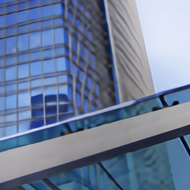} \hspace{\fsdttwofigBD} &
				
				\\
				HR~ (PSNR) \hspace{\fsdttwofigBD} &
				EDSR~\cite{lim:CVPR:2017:EDSR}~ (23.78dB) \hspace{\fsdttwofigBD} &
				DPDNN~\cite{dong:TPAMI:2018DPDNN}~ (23.49dB) \hspace{\fsdttwofigBD} &
				RDN~\cite{zhang:CVPR:2018RDN}~ (23.81dB) \hspace{\fsdttwofigBD} &
				
				\\
				\includegraphics[width=\widthscalefive \textwidth]{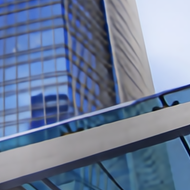} \hspace{\fsdttwofigBD} &
				\includegraphics[width=\widthscalefive \textwidth]{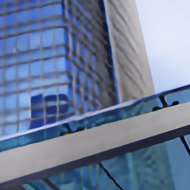} \hspace{\fsdttwofigBD} &
				\includegraphics[width=\widthscalefive \textwidth]{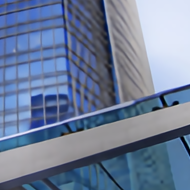} \hspace{\fsdttwofigBD} &
				\includegraphics[width=\widthscalefive \textwidth]{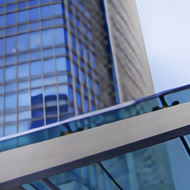} \hspace{\fsdttwofigBD} &
				
				\\
				RCAN~\cite{zhang:ECCV:2018RCAN}~ (24.00dB) \hspace{\fsdttwofigBD}&
				D-DBPN~\cite{haris2018deep}~ (23.52dB)\hspace{\fsdttwofigBD} &
				SRFBN~\cite{li:CVPR:2019feedback}~ (23.85dB)\hspace{\fsdttwofigBD} &
				\textbf{Ours~} (\textbf{24.15dB})\hspace{\fsdttwofigBD} &
				
				\\
			\end{tabular}
		\end{adjustbox}
		\vspace{0.5mm}
		
	\end{tabular}
	\caption{
		Visual quality and PSNR comparisons of different SISR methods for a sample image in the $Urban100$ dataset (bicubic-downsampling, $\times4$).}
	\label{fig:BI_urban46}
\vspace{-5mm}
\end{figure*}

\subsection{Deep Unfolding Network for NARM Model}

When compared with the image prior model in Eq. (\ref{model-based}), NARM is more sophisticated because it involves two regularization terms: one specified by the denoising prior (a variation of DPDNN \cite{dong:TPAMI:2018DPDNN}) and the other related to the nonlocal AR model \cite{dong2013sparse} (a brand-new design). Formally, we consider the following nonconvex optimization problem
\begin{equation}
\tx = \argmin_{\tx}||\ty-\tA \tx||_2^2  + \mu||\ty-\tA\tS\tx||_2^2+\lambda\tOmega(\tx) \label{problem1}
\end{equation}
where $\tOmega(\tx)$ and $\tS\tx$ denote denoising-based prior and NARM-defined prior respectively. Using Half-Quadratic Splitting \cite{he2013half}, we can rewrite the above optimization problem into
\begin{equation}
\footnotesize
\begin{split}
(\tx,\tv)  = \argmin_{\tx,\tv}||\ty-\tA \tx||_2^2 + \mu||\ty-\tA\tS\tx||_2^2+\eta||\tx-\tv||_2^2
+ \lambda\tOmega(\tv) \label{model-based-NARM}
\end{split}
\end{equation}
where $\tv$ is the auxiliary variable. It is well known that the above optimization problem can be solved by alternatively solving two subproblems related to $\tx$ and $\tv$ as shown in Eqs. \eqref{sub_problem_1} and \eqref{sub_problem_2} \cite{dong:TPAMI:2018DPDNN}.

Based on the NARM model in Eq. \eqref{NARM_eq}, we have $\tS\tx=\tx+\te$ where $\te=-\te_x$ is the modeling error (regardless of sign flip). It follows that the NARM optimization problem in Eq. (\ref{problem1}) can be translated into
\begin{equation}
\begin{split}
(\tx,\tv,\te) &= \argmin_{\tx,\tv,\te}||\ty-\tA \tx||_2^2 +\mu||\ty-\tA(\tx+\te)||_2^2\\
&+\gamma||\tS\tx-(\tx+\te)||_2^2+\eta||\tx-\tv||_2^2 + \lambda\tOmega(\tv)  \label{problem}
\end{split}
\end{equation}
As an extension of previous result \cite{dong:TPAMI:2018DPDNN}, we note that the newly-formulated NARM-based optimization problem can be solved by alternatively solving the following three sub-problems
\begin{subequations}
\begin{equation}
\begin{split}
\tx^{(t+1)} &= \argmin_{\tx}||\ty-\tA \tx||_2^2+\mu||\ty-\tA(\tx+\te^{(t)})||_2^2\\
&+\gamma||\tS\tx-(\tx+\te^{(t)})||_2^2 +\eta||\tx-\tv^{(t)}||_2^2\label{Recon_problem}
\end{split}
\end{equation}

\begin{equation}\te^{(t+1)} =\argmin_{\te}\mu||\ty-\tA(\tx^{(t)}+\te)||_2^2+\gamma||\tx^{(t)}+\te-\tS\tx^{(t)}||_2^2\label{AR_error_problem} \end{equation}

\begin{equation}\tv^{(t+1)} = \argmin_{\tv}\eta||\tx^{(t)}-\tv||_2^2+\lambda\tOmega(\tv) \label{RNN_problem} \end{equation}
\label{NARM_problem}
\end{subequations}

In traditional model-based approaches \cite{boyd2011distributed,he:CVPR:2016deep}, alternatively solving the above three equations requires many iterations to converge leading to prohibitive computational cost. Meantime, the regularization functions and the hyper-parameters cannot be jointly optimized in an end-to-end manner. To address those issues, we propose to unfold the NARM-based optimization in Eq. \eqref{NARM_problem} into a concatenation of repeating network modules as shown in Fig. \ref{fig:framework}~(a). Different with previous work DPDNN \cite{dong:TPAMI:2018DPDNN}, the proposed network architecture of our MoG-DUN incorporates two new modules (nonlocal-AR and reconstruction) in addition to the denoising module. Note that both short and long skip connections \cite{huang2017densely} are incorporated to facilitate the information flow across different stages.
Furthermore, we have shown the corresponding relationship between the Iterative algorithm implemented by Eq. \eqref{NARM_problem} and our proposed MoG-DUN network in Fig. \ref{fig:framework} (f), where the solutions~(\ref{RNN_problem},\ref{AR_error_solver},\ref{Recon_solver}) of iterative updating variables $\bm{v}^{(t+1)},\bm{e}^{(t+1)},\bm{x}^{(t+1)}$  exactly match the corresponding modules of proposed deep unfolding network. To facilitate the visual inspection, we have highlighted the core modules in our MoG-DUN by red color.

The $T$ repeating stages in MoG-DUN exactly executes $T$ iterations of Eq. \eqref{NARM_problem}. In our current implementation, a total of $T=4$ stages was adopted. Each stage of MoG-DUN consists of three basic building blocks: U-net based deep denoising module, a fast nonlocal-AR module, and a versatile reconstruction module. The deep denoising module is responsible for the updating of auxiliary variable $\tv^{(t+1)}$ as described in Eq. \eqref{NARM_problem}; the fast nonlocal-AR module calculates the NARM matrix $\tS$ as defined by Eq. \eqref{NARM_S} and the corresponding $\tS\tx^{(t)}$ in a computationally efficient manner. The versatile reconstruction module takes auxiliary variable $\tv^{(t+1)}$ and nonlocal AR model $\tS\tx^{(t)}$ as inputs and output reconstructed image $\tx^{(t+1)}$. Then, the updated $\tx^{(t+1)}$ is fed into the next stage to refine the estimate of $\tv$ and $\tS$ again. The denoising module, nonlocal-AR module and reconstruction module are alternatively updated $T$ times until reaching the final reconstruction. We will elaborate on each module next. 



\subsection{Deep Denoising Module via Dense RNN}

In general, any existing image denoising network can be used as the denoising module here. In this paper, we have adopted a variant of U-net as the backbone of our denoising module. The reasons are summarized as follows. First, the U-net has been demonstrated to have great performance in the domain of image restoration \cite{dong:TPAMI:2018DPDNN} including the task of image denoising. Second, the U-net structure gradually decreases the  spatial resolution of feature maps along the encoding process leading to reduced computation. Last, the U-net structure can exploit the multi-scale redundancies of natural images, which facilitates the recovery of spatially high-frequency information such as edges and textures. 
\begin{figure}[htbh]
\centering
\includegraphics[width=1.0\linewidth]{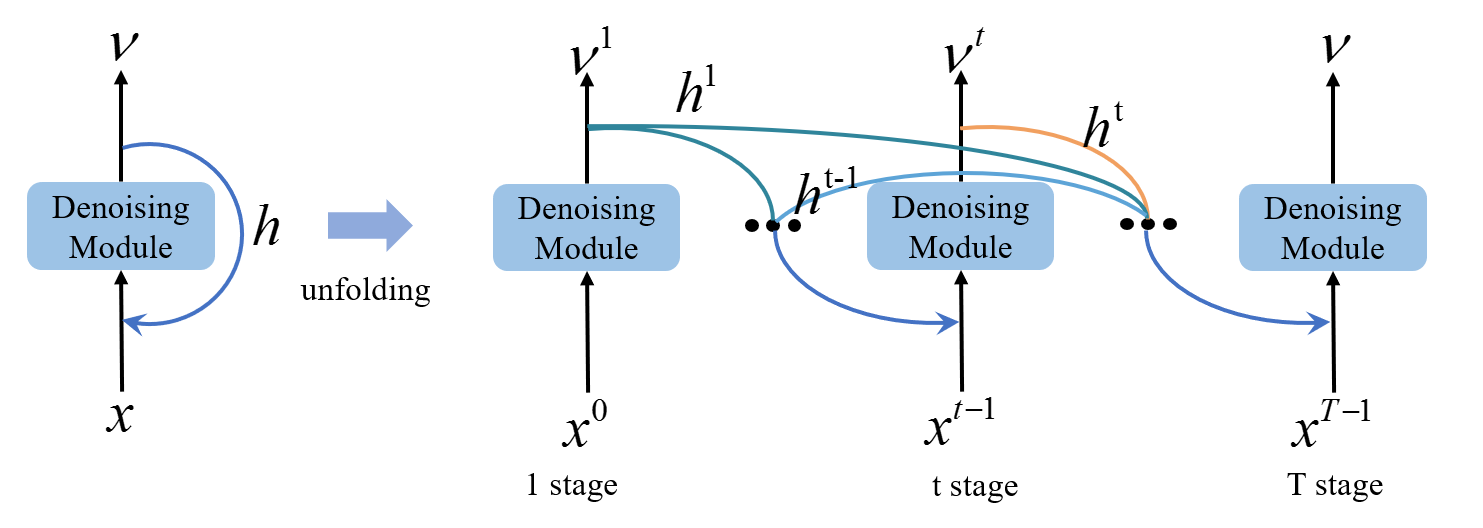}
\caption{The illustrations of the Dense RNN in the proposed network. The blue block represents deep denoising module called Dense RNN ($\tbh^{1},\tbh^{t-1},\tbh^{t}$ denote the hidden states of corresponding stages).}
  \label{fig:DRNN}
\end{figure}

\begin{figure*}[htbp]

	\scriptsize
	\centering
	\begin{tabular}{cc}
		\begin{adjustbox}{valign=t}
		\tiny
			\begin{tabular}{c}
				\includegraphics[width=0.247\textwidth]{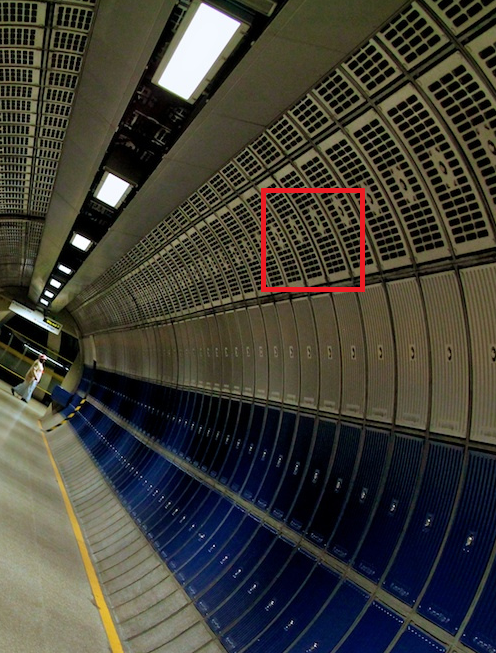}
				\\
				 \emph{Img\_078}~ from Urban100 $\times4$
				
				
			\end{tabular}
		\end{adjustbox}
		\hspace{-2.3mm}
		\begin{adjustbox}{valign=t}
		\tiny
			\begin{tabular}{cccccc}
				\includegraphics[width=0.155 \textwidth]{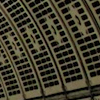} \hspace{\fsdttwofigBD} &
				\includegraphics[width=0.155 \textwidth]{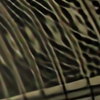} \hspace{\fsdttwofigBD} &
				\includegraphics[width=0.155 \textwidth]{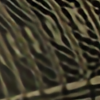} \hspace{\fsdttwofigBD} &
				\includegraphics[width=0.155 \textwidth]{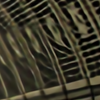} \hspace{\fsdttwofigBD} &
				
				\\
				HR~ (PSNR) \hspace{\fsdttwofigBD} &
				EDSR~\cite{lim:CVPR:2017:EDSR}~ 27.58dB)  \hspace{\fsdttwofigBD}&
				DPDNN~\cite{dong:TPAMI:2018DPDNN}~ (26.95dB) \hspace{\fsdttwofigBD} &
				RDN~\cite{zhang:CVPR:2018RDN}~ (27.82dB) \hspace{\fsdttwofigBD} &

				\\
				\includegraphics[width=0.155 \textwidth]{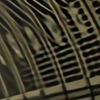} \hspace{\fsdttwofigBD} &
				\includegraphics[width=0.155 \textwidth]{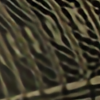} \hspace{\fsdttwofigBD} &
				\includegraphics[width=0.155 \textwidth]{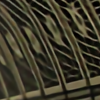} \hspace{\fsdttwofigBD} &
				\includegraphics[width=0.155 \textwidth]{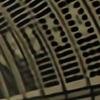} \hspace{\fsdttwofigBD} &
				
				\\
				RCAN~\cite{zhang:ECCV:2018RCAN}~ (27.92dB) \hspace{\fsdttwofigBD}&
				D-DBPN~\cite{haris2018deep}~ (26.95dB)\hspace{\fsdttwofigBD} &
				SRFBN~\cite{li:CVPR:2019feedback}~ (27.73dB)\hspace{\fsdttwofigBD} &
				\textbf{Ours~} (\textbf{28.29dB})\hspace{\fsdttwofigBD} &

				\\
			\end{tabular}
		\end{adjustbox}
		\vspace{0.5mm}
	\end{tabular}
	\caption{
	Visual quality and PSNR comparisons of different SISR methods for a sample image in the $Urban100$ dataset (bicubic-downsampling, $\times4$).}
	\label{fig:BI_urban78}
\vspace{-5mm}
\end{figure*}

Alternatively, our design of denoising module can be viewed as recurrent neural network (RNN) \cite{gregor2015draw} as shown in Fig. \ref{fig:framework}. Let $\tbh_1,...,\tbh_t$ denote the hidden states of $t$ stages, which will be used in the next stages of dense RNN. To better illustrate this particular structure, we have shown the dense RNN structure with more details in Fig. \ref{fig:DRNN}. It can be seen that  multiple hidden states are reused by the denoising module, which is equivalent to the implementation by a RNN structure.
However, different from exiting RNN methods \cite{kim:CVPR:2016DRCN,tai:CVPR:2017DRRN,li:CVPR:2019feedback} for SISR receiving only one state $\tbh^{t-1}$ of former stage, we propose to leverage multiple states $\tbh^1,...,\tbh^{t-1}$ of former stages through long connections. As shown in Fig. \ref{fig:framework}~(a), the processed information can be leveraged to refine the current image reconstruction at the $(t+2)$-th stage by receiving former states $\tbh^1,...,\tbh^{t-1},\tbh^{t},\tbh^{t+1}$ at previous $t+1$ stages \cite{zhao:IJCAI:2019recurrent}.
Exploiting the hidden states of previous stages allows us to more faithfully reconstruct the missing high-frequency information for SISR (please refer to Figs. \ref{fig:BI_urban46}-\ref{fig:BI_set14_02} for concrete image examples).

As shown in Denoising Module of Fig.~\ref{fig:framework}(a), the encoder part consists of four encoding blocks (EB) and four decoding blocks (DB).
Except the last EB, each EB is followed by a downsampling layer that sub-samples the feature maps with scaling factor of two along both axes to increase the receipt field of neurons. As shown in the left part of Fig. \ref{fig:framework}(b), each EB consists of three convolutional layers with $3\times3$ kernels, a residual layer and ReLU nonlinearity to generate 64 channel feature maps. The decoder reconstructs the image with four DBs, each of which contains three convolutional layers and a residual layer as shown in the right part of Fig. \ref{fig:framework} (b).  Except the last DB, each DB is followed by a deconvolution layer to increase the spatial size of feature maps by a scaling factor of two. To compensate the lost spatial information, upsampled feature maps are concatenated with the feature maps of the same spatial dimension from the encoder. Thanks to the transparency of our design, all dense-RNN module in the T stages share the same network parameters. 

\subsection{Fast Nonlocal-AR Module}

The nonlocal-AR module corresponds to the unfolding of NARM matrix $\tS$ into a network implementation. Based on the observation that natural images often contain rich repetitive structures, nonlocal similarity has shown effective for recovering missing high-frequency information in SISR. In model-based implementation, finding similar patches is often the computational bottleneck because nearest-neighbor search is an NP-hard problem \cite{indyk1998approximate}. By contrast, calculating the nonlocal relationship among image patches can be implemented efficiently in parallel by  nonlocal  neural networks \cite{wang:cvpr:2018non-local}. Inspired by the design of nonlocal operation in \cite{wang:cvpr:2018non-local} and its application into image restoration \cite{liu:nips:2018non-local}, we have designed a {\it fast} nonlocal operation module for computing NARM matrix $\tS$ here. Fig. \ref{fig:framework}~(c) illustrates the block diagram of implementing a non-local operation (highlighted by orange color) designed for computing the similarity for a given image. Following the formulation in nonlocal-mean filtering \cite{buades:CVPR:2005nonlocal} and bilateral filtering \cite{tomasi1998bilateral}, a non-local operation can be defined as
\begin{equation}
\ty_i = ( \sum_{\forall j}f(\tx_i,\tx_j)g(\tx_j) )/\sum_{\forall j}f(\tx_i,\tx_j)
\label{non-local operater}
\end{equation}
where $i$ is the index of an output position (e.g., in space or time), the $j$ is the enumeration of all possible positions, and the pairwise similarity function $f$ calculates the relationship between $i$ and all $j$. Similar to \cite{liu:nips:2018non-local}, we only calculate the $q\times q$ block centered at position $i$ instead of the whole image. For a balanced trade-off between cost and performance, we have chosen $q=15$.  

The design of similarity function $f$ has been considered in \cite{wang:cvpr:2018non-local}. For example, an embedded Gaussian function can be used to calculate similarity
\begin{equation}
f(\tx_i,\tx_j) = e^{(\theta(\tx_i)^T\phi(\tx_j))}
\label{non-local sub1}
\end{equation}
where $\theta(\tx_i)=W_\theta \tx_i$, $\phi(\tx_j)=W_\phi \tx_j$, and $W_\theta,W_\phi$ are the weight matrices. For simplicity, we opt to employ a linear embedding for $g(\tx_j)=W_g \tx_j$, where $W_g$ is a learnable weight matrix. Then the output of nonlocal-AR block $\tS\tx_i$ is calculated by
\begin{equation}
\tS\tx_i = \tW_\omega \ty_i + \tx_i = \tW_\omega \sigma[\theta(\tx_i)^T\phi(\tx_j)]g(\tx_j)+\tx_i
\end{equation}
where $\sigma$ denotes the $softmax$ operator and $W_\omega$ is the embedding weight matrix. Note that the pairwise computation of a non-local block enjoys the benefit of being lightweight because its computational cost implemented by matrix multiplication is comparable to a typical
convolutional layer in standard networks. Moreover, pairwise computation of  nonlocal blocks can be used in high-level, sub-sampled feature maps (e.g., using the subsampling trick as described in \cite{wang:cvpr:2018non-local}). As shown in \cite{yue2018compact}, nonlocal module admits even more efficient implementations by considering a compact representation for multiple kernel functions with Taylor expansion. 

\subsection{Versatile Reconstruction Module}
\label{recon_module}
With the output of denoising module $\tv^{(t+1)}$ and nonlocal-AR module $\tS\tx^{(t)}$, we can reconstruct the updated image $\tx^{(t+1)}$ in two steps. First, we need to calculate $\te^{(t+1)}$ by solving Eq.~(\ref{AR_error_problem}) using a single step of gradient descent
\begin{equation}\te^{(t+1)}
\footnotesize
=\te^{(t)}-\delta[\mu\tA^T(\tA(\tx^{(t)}+\te^{(t)})-\ty)+\gamma(\tx^{(t)}+\te^{(t)}-\tS\tx^{(t)})]\label{AR_error_solver} \end{equation}
where $\delta$ is a parameter controlling the step size of convergence.
Then, we can reconstruct new $\tx^{(t+1)}$ with updated $\te^{(t+1)}$ and $\tv^{(t+1)}$ by solving Eq.~(\ref{Recon_problem}) using another single step of gradient descent
\begin{equation}
\footnotesize
\begin{split}
\tx^{(t+1)}  =& \tx^{(t)}-\delta^{'}[\tA^T(\tA\tx^{(t)}-\ty)+\mu\tA^T(\tA(\tx^{(t)}-\te^{(t+1)})-\ty) \\
& +\gamma(\tx^{(t)}+\te^{(t+1)}-\tS\tx^{(t)})+\eta(\tx^{(t)}-\tv^{(t+1)})]\label{Recon_solver}
\end{split}
 \end{equation}
where $\delta^{'}$ is another relaxation parameter.

Note that Eqs.~\eqref{AR_error_solver},\eqref{Recon_solver} still involve sophisticated degradation matrix $\tA$ that is expensive to calculate. In DPDNN \cite{dong:TPAMI:2018DPDNN}, the pair of operators $\tA$ and $\tA^T$ were replaced by downsampling and upsampling operators but at the price of limited modeling capability (e.g., they can not handle multiple degradation kernels \cite{zhang:CVPR:2018SRMDNF}). Different from DPDNN, we propose a more {\it versatile} design here - i.e., to simulate the forward and inverse process of degradation by a shallow four-layer convolutional network.  Specifically, the degradation process $\tA$ is simulated by a network called {\it Down-Sampling-network (DS-net)} consisting of three convolutional layers with $3\times3$ kernels and 64 channels and one convolutional  layer to decrease the spatial resolution with corresponding scale as shown in Fig.~\ref{fig:framework}~(d). In a similar way, the network called {\it Up-Sampling-network (US-net)} representing $\tA^T$ consists of three convolutional layers with $3\times3$ kernels and 64 channels and one deconvolution layer to increase the spatial resolution with corresponding scale  as shown in Fig.~\ref{fig:framework}~(e) .

The versatility of our newly-design reconstruction module is further demonstrated by its capability of handling multiple blur kernels \cite{zhang:CVPR:2018SRMDNF}. Dealing with multiple degradations is highly desirable in practical SISR scenarios where image degradation is complex and even spatially-varying. In order to handle multiple blur kernels, one can expand the blur kernel to the same spatial dimension as the input images using a strategy called ``dimensionality stretching'' \cite{zhang:CVPR:2018SRMDNF}. Specifically, assuming the blur kernel is sized by $k\times k$, the blur kernel is first stretched into a vector of size $ k^2 \times 1$ and then projected onto a $d-$dimensional($d<k^2$) linear space by the principal component analysis (PCA) to reduce the computation. This way, the blur kernel map consisting of size $d\times H\times W$ is stretched from the original blur kernel of size $k\times k$, where $H$ and $W$ denote the height and width of corresponding variable. Along the channel dimension, we concatenate the stretched blur kernel maps with the corresponding intermediate result which needs to be upsampled by a \emph{US-net} or downsampled by a \emph{DS-net}. Then, the concatenated feature maps are fed into the corresponding \emph{US-net} or \emph{DS-net}. With the \emph{Reconstruction Module}, our model can achieve great performance in terms of handling multiple different degradations caused by varying blur kernels (please refer to Table \ref{tab:deblur}).

\begin{figure*}[htbp]
	\newlength\fsdttwofigBDS
	\setlength{\fsdttwofigBDS}{-2.0mm}
	\scriptsize
	\centering
	\begin{tabular}{cc}
		\begin{adjustbox}{valign=t}
		\tiny
			\begin{tabular}{c}
				\includegraphics[width=0.243\textwidth]{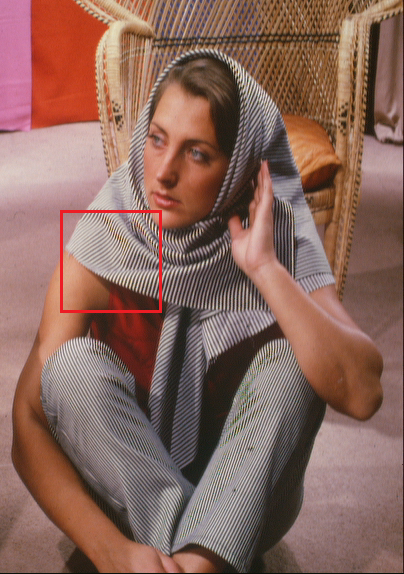}
				\\
				 \emph{Img\_002}~ from Set14 $\times2$
				
				
			\end{tabular}
		\end{adjustbox}
		\hspace{-2.3mm}
		\begin{adjustbox}{valign=t}
		\tiny
			\begin{tabular}{cccccc}
				\includegraphics[width=\widthscalefive \textwidth]{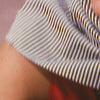} \hspace{\fsdttwofigBDS} &
				\includegraphics[width=\widthscalefive \textwidth]{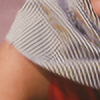} \hspace{\fsdttwofigBDS} &
				\includegraphics[width=\widthscalefive \textwidth]{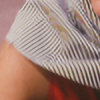} \hspace{\fsdttwofigBDS} &
				\includegraphics[width=\widthscalefive \textwidth]{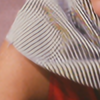} \hspace{\fsdttwofigBDS} &
				
				\\
				HR~ (PSNR) \hspace{\fsdttwofigBDS} &
				EDSR~\cite{lim:CVPR:2017:EDSR}~ (28.38dB) \hspace{\fsdttwofigBDS} &
				DPDNN~\cite{dong:TPAMI:2018DPDNN}~ (28.21dB) \hspace{\fsdttwofigBDS} &
				RDN~\cite{zhang:CVPR:2018RDN}~ (28.38dB) \hspace{\fsdttwofigBDS} &

				\\
				\includegraphics[width=\widthscalefive \textwidth]{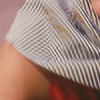} \hspace{\fsdttwofigBDS} &
				\includegraphics[width=\widthscalefive \textwidth]{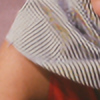} \hspace{\fsdttwofigBDS} &
				\includegraphics[width=\widthscalefive \textwidth]{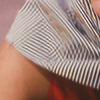} \hspace{\fsdttwofigBDS} &
				\includegraphics[width=\widthscalefive \textwidth]{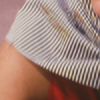} \hspace{\fsdttwofigBDS} &
				
				\\
				RCAN~\cite{zhang:ECCV:2018RCAN}~ (28.44dB) \hspace{\fsdttwofigBDS}&
				D-DBPN~\cite{haris2018deep}~ (28.40dB)\hspace{\fsdttwofigBDS} &
				SRFBN~\cite{li:CVPR:2019feedback}~ (29.27dB)\hspace{\fsdttwofigBDS} &
				\textbf{Ours~} (\textbf{30.16dB})\hspace{\fsdttwofigBDS} &

				\\
			\end{tabular}
		\end{adjustbox}
		\vspace{0.5mm}
		
	\end{tabular}
	\caption{
	Visual quality and PSNR comparisons of different SISR methods for a sample image in the $Set14$ dataset (bicubic-downsampling, $\times2$).
	}
	\label{fig:BI_set14_02}
\vspace{-5mm}
\end{figure*}

\section{Experimental Results}
\label{sec:4}

\subsection{Experimental Settings}

\textbf{Benchmark Datasets and Performance Metrics.} We have used 800 high-quality (2K resolution) images from the DIV2K dataset \cite{timofte:CVPRW:2017DIV2K} for training. Following  \cite{dong:ECCV:2014learning,li:CVPR:2019feedback,lim:CVPR:2017:EDSR,zhang:CVPR:2018RDN,zhang:CVPR:2018SRMDNF}, five standard benchmark datasets: Set5 \cite{bevilacqua:BMCV:2012Set5}, Set14 \cite{zeyde:CCS:2010Set14}, BSD100\cite{martin:ICCV:2001BSD100}, Urban100 \cite{huang:CVPR:2015Urban100}, Manga109 \cite{matsui:MTA:2017manga}are used for testing. Performance evaluation in terms of PSNR and SSIM \cite{wang:TIP:2004SSIM} metrics is conducted on the luminance (Y) channel only.

\textbf{Degradation Models.}
In order to demonstrate the robustness of our model in varying degradation scenarios, we have designed the following experiments with different parameter settings with the degradation model.
\begin{itemize}
    \item Default setting. This is the scenario considered in most previous SISR studies - i.e., the low-resolution (LR) image is obtained by bicubic downsampling of the high-resolution (HR) image. The downsampling ratio is usually a small positive integer ($\times2$,$\times3$,$\times4$).
    \item Interpolation setting. This is the scenario consistent with the NARM study
    \cite{dong2013sparse} in which a LR image is directly down-sampled from the HR image without any anti-aliasing filtering involved. Due to the presence of aliasing, this scenario is generally believed to be more difficult than the default setting.
    \item Realistic setting. To more faithfully characterize the degradation in the real world, this setting aims at simulating multiple degradation situations caused by different Gaussian kernels \cite{zhang:CVPR:2018SRMDNF}. Similar to SRMDNF \cite{zhang:CVPR:2018SRMDNF}, we have obtained a single trained network through reconstruction module taking multiple degradation kernels. The set of degradation kernels include isotropic Gaussian blur kernels maps whose width ranges are set to [0.2, 3], [0.2, 3] and [0.2, 4] with scale factor $\times2$, $\times3$ and $\times4$ respectively. The low-resolution (LR) image $\ty$ is obtained by $\ty = (\tx\otimes k)\downarrow$, where $\tx$ represents HR image, $\otimes$ represents the convolution operator, $k$ the blur kernel and $\downarrow$ the downsampling operator. The kernel width is uniformly sampled in the above ranges and the kernel size is fixed to $21\times21$ and the projected $d-$dimensional linear space is set to $6,8,10$ with scale factor $\times2$, $\times3$, $\times4$ respectively. The mean values of kernel width are $0.5,1.3,2.6$ as shown in Table \ref{tab:deblur}. 
\end{itemize}

\textbf{Training Setting.}
Thanks to the parameter sharing across $T$ stages, the overall MoG-DUN can be trained in an end-to-end manner. In order to further reduce the number of parameters and avoid over-fitting, we enforce deep denoising module (dense-RNN) and reconstruction module to share the same parameters. Unlike DPDNN \cite{dong:TPAMI:2018DPDNN} adopting MSE loss, we have found $L_1$ loss function works better for training the proposed MoG-DUN (e.g., it can facilitate the recovery of more high-frequency information). The $L_1$-based loss function can be expressed as:

\begin{equation}
\Theta = \argmin_\theta\sum^N_{i=1}\|\mathcal{F}(\ty_i,k;\Theta)-\tx_i\|_1,
\label{loss_function}
\end{equation}
where $\ty_i$ and $\tx_i$ denote the $i$-th pair of degraded and original image patches respectively, $k$ denotes the \emph{expanded blur kernel maps} (as explained in Section \ref{recon_module}) when dealing with multiple degradations,  and $\mathcal{F}(\ty_i,k;\Theta)$ denotes the reconstructed image patch by the network with the parameter set $\Theta$. We randomly select 16 RGB LR patches sized by $48\times48$ as the inputs and stretch the blur kernels when dealing with multiple degradations (called ``dimensionality stretching'' in \cite{zhang:CVPR:2018SRMDNF}). The image patches are randomly rotated by 90\degree, 180\degree, 270\degree and flipped horizontally as standard data augmentation techniques do. The ADAM algorithm  \cite{kingma2014adam} with $\beta_1=0.9, \beta_2=0.999, \epsilon=10^{-8}$ is adopted to optimize the network. The initial learning rate is $10^{-4}$ and decreases by half for every 300 epochs. Our network is implemented under the Pytorch framework and the training time takes less than 2days using  4 NVIDIA 1080Ti GPUs.

 \begin{table*}[htbh]
\begin{center}
\centering{
\caption{Average PSNR and SSIM results for \textbf{bicubic} downsampling degradation on five benchmark datasets.
 The best performance is shown in bold and the second best performance is shown in underline.} 
\label{tab:bicubic}
\resizebox{1.8\columnwidth}{!}{
\begin{tabular}{|l|c|c|c|c|c|c|c|c|c|c|c|}
\hline
\multirow{2}{*}{Method} & \multirow{2}{*}{Scale} & \multicolumn{2}{c|}{Set5 \cite{bevilacqua:BMCV:2012Set5}} & \multicolumn{2}{c|}{Set14 \cite{zeyde:CCS:2010Set14}} & \multicolumn{2}{c|}{BSD100 \cite{martin:ICCV:2001BSD100}}
& \multicolumn{2}{c|}{Urban100 \cite{huang:CVPR:2015Urban100}} & \multicolumn{2}{c|}{Manga109 \cite{matsui:MTA:2017manga}} \\ \cline{3-12}
\multicolumn{1}{|c|}{}                        &
                                                                    & PSNR  & SSIM   & PSNR  & SSIM   & PSNR  & SSIM   & PSNR  & SSIM   & PSNR  & SSIM          \\
                                                                    \hline \hline
  EDSR\cite{lim:CVPR:2017:EDSR}          & $\times$2                       &38.11  &0.9602  &33.92  &0.9195  &32.32  &0.9013  &32.93  &0.9351  &39.10  &0.9773               \\
  DPDNN\cite{dong:TPAMI:2018DPDNN}       & $\times$2                       &37.75  &0.9600  &33.30  &0.9150  &32.09  &0.8990  &31.50  &0.9220  &  -    &   -            \\
  DSRN \cite{han:CVPR:2018DSRN}          & $\times$2                       &37.66  &0.9590  &33.15  &0.9130  &32.10  &0.8970  &30.97  &0.9160  &  -    &   -            \\
  RDN\cite{zhang:CVPR:2018RDN}           & $\times$2                       &38.24  &\textbf{0.9614}  &\underline{34.01}  &\underline{0.9212}  &32.34  &0.9017  &\underline{32.89}  &0.9353  &39.18  &0.9780           \\
  RCAN \cite{zhang:ECCV:2018RCAN}        & $\times$2                       &\textbf{38.27}  &\textbf{0.9614}  &\textbf{34.12}  &\textbf{0.9216}  &\textbf{32.41}  &\textbf{0.9027}  &\textbf{33.34}  &\underline{0.9384}  &\textbf{39.44}  &\textbf{0.9786} \\
  D-DBPN\cite{haris2018deep}        & $\times$2                       &38.09  &0.9600  &33.85  &0.9190  &32.27  &0.9000  &32.55  &0.9324  &38.89  &0.9775           \\
  SRMDNF\cite{zhang:CVPR:2018SRMDNF}     & $\times$2                       &37.79  &0.9601  &33.32  &0.9159  &32.05  &0.8985  &31.33  &0.9204  &38.07  &0.9761               \\
  SRFBN\cite{li:CVPR:2019feedback}       & $\times$2                       &38.11  &\underline{0.9609}  &33.82  &0.9196  &32.29  &0.9010  &32.62  &0.9328  &39.09  &0.9779            \\
USRNet\cite{zhang:CVPR:2020deepUSRNet}       & $\times$2                       &37.72  &-  &33.49  &-  &32.10  &-  &31.79  &-  &-  &-            \\
  \textbf{MoG-DUN(ours)}                                   & $\times$2                       &\underline{38.25}  &\textbf{0.9614}  &34.00  &0.9205  &\underline{32.37}  &\underline{0.9020}  &32.75  &\textbf{0.9421}  &\underline{39.37}  &\underline{0.9783}               \\ 
                                         \hline \hline
  EDSR\cite{lim:CVPR:2017:EDSR}          & $\times$3                       &34.65  &0.9280  &30.52 &0.8462  &29.25  &0.8093  &28.80  &0.8653  &34.17  &0.9476               \\
  DPDNN\cite{dong:TPAMI:2018DPDNN}       & $\times$3                       &33.93  &0.9240  &30.02 &0.8360  &29.00  &0.8010  &27.61  &0.8420  &  -    &   -           \\
  DSRN \cite{han:CVPR:2018DSRN}          & $\times$3                       &33.88  &0.9220  &30.26 &0.8370  &28.81  &0.7970  &27.16  &0.8280  &  -    &   -           \\
  RDN\cite{zhang:CVPR:2018RDN}           & $\times$3                       &34.71  &0.9296  &30.57 &0.8468  &\underline{29.26}  &0.8093  &28.80  &\underline{0.8653}  &34.13  &0.9484           \\
  RCAN \cite{zhang:ECCV:2018RCAN}        & $\times$3                       &\underline{34.74}  &\underline{0.9299}  &\textbf{30.65} &\textbf{0.8482}  &\textbf{29.32}  &\textbf{0.8111}  &\textbf{29.09}  &\textbf{0.8702}  &\textbf{34.44}  &\textbf{0.9499}          \\
  SRMDNF\cite{zhang:CVPR:2018SRMDNF}     & $\times$3                       &34.12  &0.9254  &30.04 &0.8382  &28.97  &0.8025  &27.57  &0.8398  &33.00  &0.9403               \\
  SRFBN\cite{li:CVPR:2019feedback}       & $\times$3                       &34.70  &0.9292  &30.51 &0.8461  &28.81  &0.7868  &28.73  &0.8641  &34.18  &0.9481            \\
  USRNet\cite{zhang:CVPR:2020deepUSRNet}       & $\times$3                       &34.45  &-  &30.51  &-  &29.18  &-  &28.38  &-  &-  &-            \\
  \textbf{MoG-DUN(ours)}                                   & $\times$3                       &\textbf{34.76}  &\textbf{0.9300}  &\underline{30.63}  &\underline{0.8479}  &29.24  &\underline{0.8094}  &\underline{28.82}  &0.8651  &\underline{34.34}  &\underline{0.9490}               \\
                                         \hline \hline
  EDSR\cite{lim:CVPR:2017:EDSR}          & $\times$4                       &32.46  &0.8968  &28.80 &0.7876  &27.71  &0.7420  &26.64  &0.8033  &31.02  &0.9148              \\
  DPDNN\cite{dong:TPAMI:2018DPDNN}       & $\times$4                       &31.72  &0.8890  &28.28 &0.7730  &27.44  &0.7290  &25.53  &0.7680  &  -    &  -             \\
  DSRN \cite{han:CVPR:2018DSRN}          & $\times$4                       &31.40  &0.8830  &28.07 &0.7700  &27.25  &0.7240  &25.08  &0.7470  &  -    &   -           \\
  RDN\cite{zhang:CVPR:2018RDN}           & $\times$4                       &32.47  &0.8990  &28.81 &0.7871  &\underline{27.72}  &\underline{0.7419}  &26.61  &\underline{0.8028}  &31.00  &0.9151           \\
  RCAN \cite{zhang:ECCV:2018RCAN}        & $\times$4                       &\textbf{32.63}  &\textbf{0.9002}  &\textbf{28.87} &\textbf{0.7889}  &\textbf{27.77}  &\textbf{0.7436}  &\textbf{26.82}  &\textbf{0.8087}  &\underline{31.22}  &\textbf{0.9173}        \\
  D-DBPN\cite{haris2018deep}        & $\times$4                       &32.47  &0.8990  &28.82 &0.7860  &27.72  &0.7400  &26.38  &0.7946  &30.91  &0.9137           \\
  SRMDNF\cite{zhang:CVPR:2018SRMDNF}     & $\times$4                       &31.96  &0.8925  &28.35 &0.7787  &27.49  &0.7337  &25.68  &0.7731  &30.09  &0.9024               \\
  SRFBN\cite{li:CVPR:2019feedback}       & $\times$4                       &32.47  &0.8983  &28.81 &0.7868  &27.72  &0.7409  &26.60  &0.8015  &31.15  &0.9160            \\
  USRNet\cite{zhang:CVPR:2020deepUSRNet}       & $\times$4                       &32.45  &-  &28.83  &-  &27.69  &-  &26.44  &-  &-  &-            \\
  \textbf{MoG-DUN(ours)}                                   & $\times$4                       &\underline{32.60}  &\underline{0.8998}  &\underline{28.84}  &\textbf{0.7873}  &27.70  &0.7403  &\underline{26.63}  &0.8016  &\textbf{31.26}  &\underline{0.9169}               \\
                                         \hline

\end{tabular}
}}
\end{center}
\end{table*}

\subsection{Ablation Study}
To further verify the effectiveness of nonlocal-AR module, we have conducted an ablation study to compare the PSNR performance of MoG-DUN with and without nonlocal-AR module. In our ablation study, we have used directly downsampling degradation with different Gaussian kernel size of $0.5,1.0$ and $\times3$ directly downsampling on four frequently-used benchmark datasets. As shown in Tab.~\ref{tab:abl_nonlocal}, the nonlocal-AR module does make a contribution to the overall performance of MoG-DUN. 
\begin{table}[htbh]
\begin{center}
\centering{
\caption{Average PSNR results with and without nonlocal-AR module  for directly downsampling degradation $\times3$ with different Gaussian kernels width on four frequently-used benchmark datasets. }
\label{tab:abl_nonlocal}
\resizebox{1.0\columnwidth}{!}{
\begin{tabular}{|l|c|c|c|c|c|}
\hline
Methods                &Kernel Width   &Set5      &Set14    &BSD68  &Urban100   \\  \hline
w/o NL-AR   &0.5         &32.86      &28.99    & 28.00   &27.27    \\ \hline
w  NL-AR    &0.5            &32.98    &29.11    &28.12   &27.51    \\ \hline
w/o NL-AR   &1.0         &34.15      &30.15    & 28.90   &28.37    \\ \hline
w  NL-AR    &1.0            &34.34    &30.23    &28.99  & 28.63    \\ \hline
\end{tabular}
}}
\end{center}
\end{table}

To investigate the influence of those dense connections as shown in Fig.~\ref{fig:framework}~(a), we have conducted an ablation study to compare the PSNR performance of MoG-DUN with and without dense connections for directly downsampling degradation at the scaling factor of $\times3$. As shown in  Tab.~\ref{tab:abl_dense_connections}, we see that dense connections contribute to the PSNR gain of about $0.16~dB$ on the average.
\begin{table}[htbh]
\begin{center}
\centering{
\caption{Average PSNR results with and without dense connections of denosing module  for directly downsampling degradation $\times3$. }
\label{tab:abl_dense_connections}
\resizebox{1.0\columnwidth}{!}{
\begin{tabular}{|l|c|c|c|c|}
\hline
Methods      &Set5               &Set14                 &BSD68 &Urban100  \\  \hline
w/o dense connections              &32.19      &28.10    &27.53  & 26.92     \\ \hline
w  dense connections               &32.36      &28.21    &27.64   &27.15     \\ \hline

\end{tabular}
}}
\end{center}
\end{table}

\begin{figure}[htbh]
\centering
\includegraphics[width=1.0\linewidth]{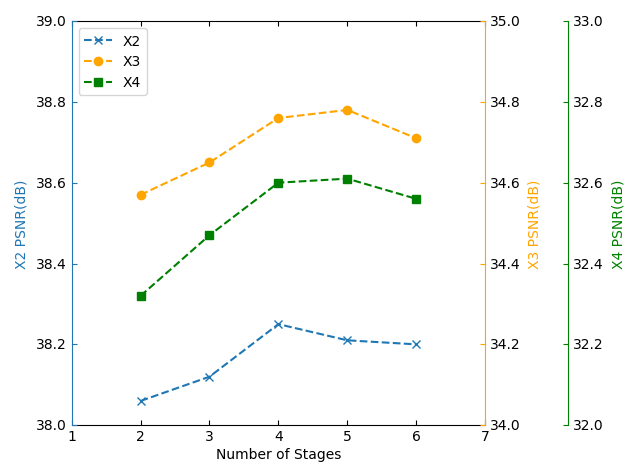}
\caption{The average PSNR performance as a function of parameter $T$ (the total number of Unet stages) of proposed MoG-DUN with $\times2,\times3,\times4$ bicubic-downsampling on Set5 \cite{bevilacqua:BMCV:2012Set5}.}
  \label{fig:psnr_stages}
\end{figure}

To explore the impact of the number of unfolded stages on the SISR performance, we have conducted another experiment with varying the parameter $T$. Fig.~\ref{fig:psnr_stages} shows the average PSNR results of different stages $T$ from two to six with $\times2$,$\times3$ and $\times4$ bicubic-downsampling. It can be seen that the PSNR increases as the number of stages increases. However, the PSNR improvement rapidly saturates when $T \geqslant 4$, which justifies the choice of $T = 4$ in our implementation to balance the performance and computational complexity.

 We have conducted experiments with the case where the parameters are not shared  in three ($\times2,\times3,\times4$) bicubic-downsampling settings when $T=4$. As shown in Table \ref{tab:abl_parameter_share}, disabling parameter sharing does not lead to any noticeable performance gain at the price of quadrupled number of parameters. Based on such experimental finding, we conclude that parameter sharing is a good strategy for MoG-DUN. It is also worth mentioning that parameter sharing has also been widely considered as an effective strategy of shrinking the optimization gap in neural architecture search \cite{xie2020weight}.
 \begin{table}[htbh]
\begin{center}
\centering{
\caption{Average PSNR results with and without parameters sharing of denosing module for bicubic downsampling degradation with $\times2,\times3,\times4$ on Set5 datasets. }
\label{tab:abl_parameter_share}
\resizebox{0.8\columnwidth}{!}{
\begin{tabular}{|l|c|c|c|c|}
\hline
Methods      &$\times2$               & $\times3$               &$\times4$   \\  \hline
w/o params sharing              &38.27      &34.79    &32.63       \\ \hline
w  params sharing               &38.25      &34.76    &32.60        \\ \hline

\end{tabular}
}}
\end{center}
\end{table}

\subsection{Experimental Results for the Default Setting}
For bicubic downsampling, we have compared MoG-DUN with eight state-of-the-art image SR methods: EDSR \cite{lim:CVPR:2017:EDSR}, DPDNN \cite{dong:TPAMI:2018DPDNN}, DSRN \cite{han:CVPR:2018DSRN}, RDN \cite{zhang:CVPR:2018RDN}, RCAN \cite{zhang:ECCV:2018RCAN}, D-DBPN \cite{haris2018deep}, SRMDNF \cite{zhang:CVPR:2018SRMDNF}, SRFBN \cite{li:CVPR:2019feedback}, USRNet \cite{zhang:CVPR:2020deepUSRNet}. 
The average PSNR and SSIM results of eight benchmark methods in Tab.\ref{tab:bicubic} are cited from corresponding papers. It is easy to see that our method is superior to most of competing methods in terms of PSNR and SSIM values. When compared with a much deeper network RCAN \cite{zhang:ECCV:2018RCAN} involving over 400 convolutional layers, we can achieve highly comparable and sometimes even better results.

\begin{figure*}[htbp]
	\scriptsize
	\centering
    
		\begin{adjustbox}{valign=t}
		\tiny
			\begin{tabular}{cccccc}
				\includegraphics[width= 0.16\textwidth]{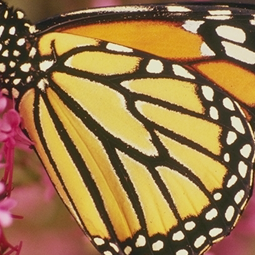} \hspace{-3mm} &
                \includegraphics[width=0.16 \textwidth]{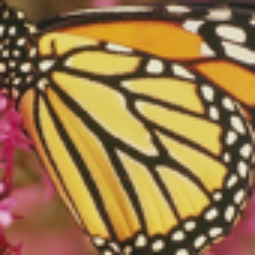} \hspace{-3mm} &
				\includegraphics[width=0.16 \textwidth]{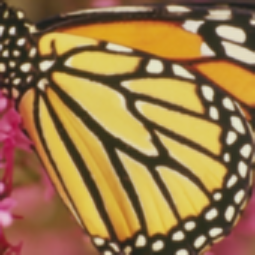} \hspace{-3mm} &
				\includegraphics[width=0.16 \textwidth]{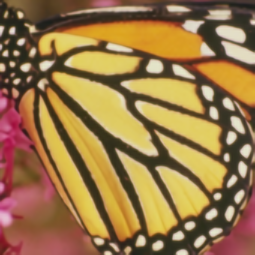} \hspace{-3mm} &
				\includegraphics[width=0.16 \textwidth]{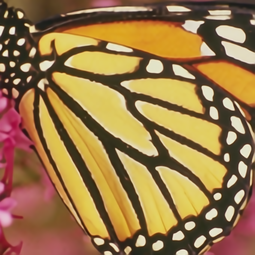} \hspace{-3mm} &
				
				\\
                
				HR~  \hspace{-3mm} &
				T=1~ \hspace{-3mm} &
				T=2~ \hspace{-3mm} &
			    T=3~ \hspace{-3mm} &
			    T=4~ \hspace{-3mm} &
		
				\\
			\end{tabular}
		\end{adjustbox}
		\vspace{0.5mm}
		
	\caption{
		SISR intermediate visual results of different stages on `Img\_003' from Set5 \cite{huang:CVPR:2015Urban100} (bicubic-downsampling, $\times3$).
	}
	\label{fig:intermediate_results}
\end{figure*}

The image comparison results for a scale factor of $\times 4$ are reported in Fig.~\ref{fig:BI_urban46}. For this specific example, our SR-resolved result of `Img\_046' from $Urban100$ are recovered with fewer visible artifacts (e.g., the glassy surface on the right side of the building) than other competing methods. Note that our PSNR result is also noticeably higher than the previous state-of-the-art RCAN at a lower cost. In another  challenging example (`Img\_078' from $Urban100$ dataset), our method can recover much more faithful textured details as shown in Fig.\ref{fig:BI_urban78}; while all other competing methods suffer from severe aliasing artifacts (i.e., distorted tile patterns). The visual quality improvement achieved by MoG-DUN is mainly due to the fact that our proposed model makes full use of the feature maps from former stages to refine the final result. Taking one more classic example known for its notorious aliasing distortion, Fig.~\ref{fig:BI_set14_02} shows the results of bicubic $\times 2$ degradation of `Img\_002' from the $Set14$ dataset. 
Additionally, we have shown the intermediate image comparison results of different stages in Fig.~\ref{fig:intermediate_results}, from which we can see that more high-frequency information has been recovered along with the increasing number of stages.

\vspace{3mm}

\subsection{Experimental Results for the Interpolation Setting}

\begin{figure*}[htbp]
	\scriptsize
	\centering
	\newlength\fsdurthree
	\setlength{\fsdurthree}{-3.2mm}
		\begin{adjustbox}{valign=t}
		\tiny
			\begin{tabular}{ccccccccc}
				\includegraphics[width=0.130 \textwidth]{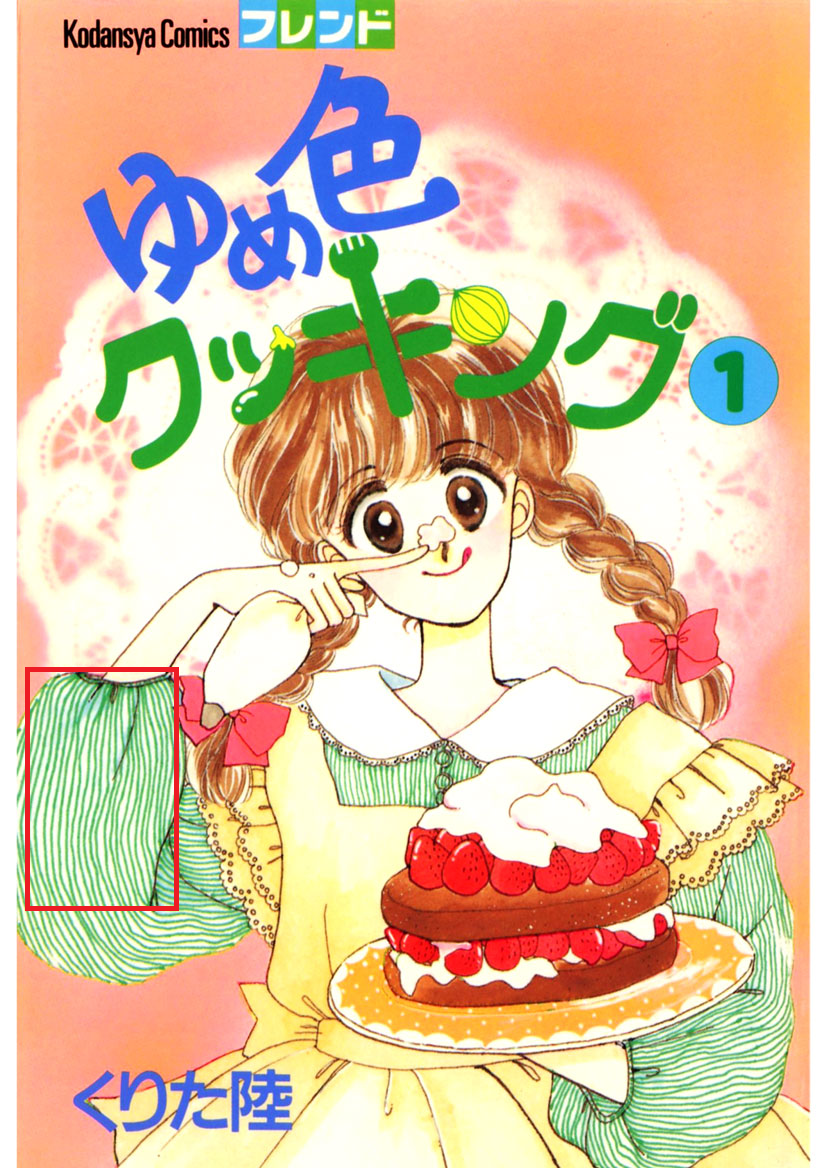} \hspace{\fsdurthree} &
                \includegraphics[width=\widthscaleeight \textwidth]{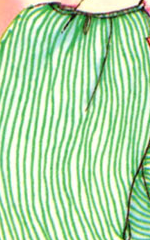} \hspace{\fsdurthree} &
				\includegraphics[width=\widthscaleeight \textwidth]{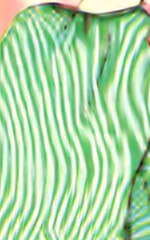} \hspace{\fsdurthree} &
				\includegraphics[width=\widthscaleeight \textwidth]{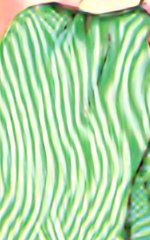} \hspace{\fsdurthree} &
				\includegraphics[width=\widthscaleeight \textwidth]{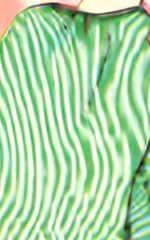} \hspace{\fsdurthree} &

				\includegraphics[width=\widthscaleeight \textwidth]{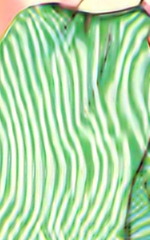} \hspace{\fsdurthree} &
				\includegraphics[width=\widthscaleeight \textwidth]{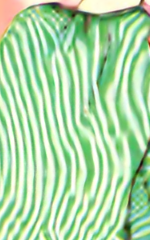} \hspace{\fsdurthree} &
				\includegraphics[width=\widthscaleeight \textwidth]{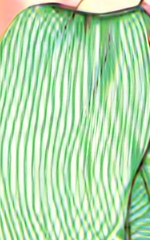} \hspace{\fsdurthree} &
				
				\\
                \emph{Img\_109}~ from Manga109&
				HR~ (PSNR) \hspace{\fsdurthree} &
				EDSR~\cite{lim:CVPR:2017:EDSR}~ (24.55dB) \hspace{\fsdurthree} &
				DPDNN~\cite{dong:TPAMI:2018DPDNN}~ (24.49dB) \hspace{\fsdurthree} &
				RDN~\cite{zhang:CVPR:2018RDN}~ (24.65dB) \hspace{\fsdurthree} &
                RCAN~\cite{zhang:ECCV:2018RCAN}~ (24.76dB)\hspace{\fsdurthree} &
				SRFBN~\cite{li:CVPR:2019feedback}~ (24.46dB)\hspace{\fsdurthree} &
				\textbf{Ours~} (\textbf{26.17dB})\hspace{\fsdurthree} &
		
				\\
			\end{tabular}
		\end{adjustbox}
		\vspace{0.5mm}
		
	\caption{
		SISR visual quality comparisons of different methods on `Img\_109' from Manga109 \cite{matsui:MTA:2017manga}. The degradation is directly downsamping with scale factor $\times4$.
	}
	\label{fig:D_manga109}
\vspace{-5mm}
\end{figure*}

For directly downsampling degradation, we have compared our model with five state-of-the-art image SR methods: EDSR \cite{lim:CVPR:2017:EDSR}, DPDNN \cite{dong:TPAMI:2018DPDNN}, RDN \cite{zhang:CVPR:2018RDN}, RCAN \cite{zhang:ECCV:2018RCAN},  SRFBN \cite{li:CVPR:2019feedback}.
The average PSNR and SSIM results for three scaling factors of $\times2,\times3,\times4$ are shown in Tab.\ref{tab:directly}. The PSNR/SSIM results of five benchmark methods are retrained from the original source codes released by their authors. From Tab.\ref{tab:directly}, we can observe that the proposed method is consistently superior to all five benchmark methods in terms of both PSNR and SSIM values. Subjective quality comparison results for a cartoon image at the scale factor of $\times 4$ are shown in Fig.\ref{fig:D_manga109}. Apparently, our SR result of `Img\_109' is the closest to that of the ground-truth both subjectively and objectively; the PSNR gain over other competing methods is over $1.4dB$ for this specific case.

\begin{table*}[htbh]
\begin{center}
\centering{
\caption{Average PSNR and SSIM results for \textbf{directly}  downsampling degradation on five benchmark datasets.
 The best performance is shown in bold.}
\label{tab:directly}
\resizebox{1.8\columnwidth}{!}{
\begin{tabular}{|l|c|c|c|c|c|c|c|c|c|c|c|}
\hline
\multirow{2}{*}{Method} & \multirow{2}{*}{Scale} & \multicolumn{2}{c|}{Set5 \cite{bevilacqua:BMCV:2012Set5}} & \multicolumn{2}{c|}{Set14 \cite{zeyde:CCS:2010Set14}} & \multicolumn{2}{c|}{BSD100 \cite{martin:ICCV:2001BSD100}}
& \multicolumn{2}{c|}{Urban100 \cite{huang:CVPR:2015Urban100}} & \multicolumn{2}{c|}{Manga109 \cite{matsui:MTA:2017manga}} \\ \cline{3-12}
\multicolumn{1}{|c|}{}                        &
                                                                    & PSNR  & SSIM   & PSNR  & SSIM   & PSNR  & SSIM   & PSNR  & SSIM   & PSNR  & SSIM          \\
                                                                    \hline \hline

  EDSR\cite{lim:CVPR:2017:EDSR}          & $\times$2                       &35.57  &0.9444  &31.30  &0.8864  &30.21  &0.8666  &28.82  &0.8971  &35.25  &0.9446               \\
  DPDNN\cite{dong:TPAMI:2018DPDNN}       & $\times$2                       &35.51  &0.9445  &31.26  &0.8865  &30.13  &0.8648  &28.79  &0.8972  &35.15  &0.9641          \\
  RDN\cite{zhang:CVPR:2018RDN}           & $\times$2                       &35.77  &0.9458  &31.38  &0.8870  &30.36  &0.8692  &29.54  &0.9071  &35.67  &0.9658           \\
  RCAN \cite{zhang:ECCV:2018RCAN}        & $\times$2                       &35.63  &0.9449  &31.34  &0.8867  &30.30  &0.8677  &29.19  &0.9024  &35.59  &0.9651         \\
  SRFBN\cite{li:CVPR:2019feedback}       & $\times$2                       &35.66  &0.9450  &31.35  &0.8866  &30.28  &0.8670  &29.04  &0.9002  &35.42  &0.9648            \\
  \textbf{MoG-DUN(ours)}                                   & $\times$2                       &\textbf{35.99}  &\textbf{0.9474}  &\textbf{31.77}  &\textbf{0.9333}  &\textbf{30.53}  &\textbf{0.8724}  &\textbf{30.44}  &\textbf{0.9200}  &\textbf{36.33}  &\textbf{0.9689}                \\

                                         \hline \hline

  EDSR\cite{lim:CVPR:2017:EDSR}          & $\times$3                       &31.50  &0.8992  &27.79 &0.7976  &27.19  &0.7651  &25.41  &0.8038  &29.13  &0.9109               \\
  DPDNN\cite{dong:TPAMI:2018DPDNN}       & $\times$3                       &31.56  &0.9004  &27.82 &0.7991  &27.13  &0.7638  &25.47  &0.8057  &29.23  &0.9121          \\

  RDN\cite{zhang:CVPR:2018RDN}           & $\times$3                       &31.91  &0.9039  &28.00 &0.8017  &27.38  &0.7716  &26.02  &0.8204  &29.91  &0.9189           \\
  RCAN \cite{zhang:ECCV:2018RCAN}        & $\times$3                       &31.81  &0.9026  &27.93 &0.7994  &27.33  &0.7677  &25.98  &0.8188  &29.68  &0.9165          \\

  SRFBN\cite{li:CVPR:2019feedback}       & $\times$3                       &31.89  &0.9033  &27.96 &0.8012  &27.32  &0.7686  &25.85  &0.8174  &29.66  &0.9172            \\
  \textbf{MoG-DUN(ours)}                                   & $\times$3                       &\textbf{32.36}  &\textbf{0.9095}  &\textbf{28.21} &\textbf{0.8059}  &\textbf{27.64}  &\textbf{0.7806}  &\textbf{27.15}  &\textbf{0.8483}  &\textbf{30.65}  &\textbf{0.9279}                \\

                                         \hline \hline

  EDSR\cite{lim:CVPR:2017:EDSR}          & $\times$4                       &28.94  &0.8528  &25.98 &0.7326  &25.66  &0.6961  &23.47  &0.7303  &26.04  &0.8567             \\
  DPDNN\cite{dong:TPAMI:2018DPDNN}       & $\times$4                       &28.94  &0.8539  &25.97 &0.7341  &25.57  &0.6937  &23.44  &0.7309  &25.96  &0.8585          \\

  RDN\cite{zhang:CVPR:2018RDN}           & $\times$4                       &29.24  &0.8608  &26.16 &0.7383  &25.86  &0.7053  &24.05  &0.7527  &26.59  &0.8688           \\
  RCAN \cite{zhang:ECCV:2018RCAN}        & $\times$4                       &29.13  &0.8580  &26.18 &0.7374  &25.82  &0.7002  &23.95  &0.7486  &26.35  &0.8643        \\
  SRFBN\cite{li:CVPR:2019feedback}       & $\times$4                       &29.09  &0.8583  &26.07 &0.7354  &25.72  &0.6996  &23.71  &0.7402  &26.32  &0.8648            \\
  \textbf{MoG-DUN(ours)}                                   & $\times$4                       &\textbf{29.65}  &\textbf{0.8706}  &\textbf{26.35} &\textbf{0.7448}  &\textbf{25.98}  &\textbf{0.7107}  &\textbf{24.70}  &\textbf{0.7744}  &\textbf{26.99}  &\textbf{0.8776}               \\

                                         \hline

\end{tabular}
}}
\end{center}
\end{table*}
\vspace{-5mm}

\subsection{Experimental Results for the Realistic Setting}
This scenario is arguable more challenging than the previous two due to the uncertainty with multiple degradation kernels.  Five state-of-the-arts methods: EDSR \cite{lim:CVPR:2017:EDSR}, SRMDNF \cite{zhang:CVPR:2018SRMDNF}, RDN\cite{zhang:CVPR:2018RDN}, RCAN \cite{zhang:ECCV:2018RCAN}, SRFBN \cite{li:CVPR:2019feedback} are used to demonstrate the effectiveness of our model. For this new degradation assumption, we have to retrained their models by either released source codes or published papers to generate the final results. The benchmark methods EDSR,RDN,RCAN and SRFBN only take the LR images as inputs and lack the ability of handling multiple degradation.  For a fair comparison, we have transformed those method into handling multiple degradations by taking LR images concatenated with expanded blur kernels maps as the inputs, following SRMDNF\cite{zhang:CVPR:2018SRMDNF}.

\begin{figure*}[htbp]
	\scriptsize
	\centering
    \newlength\fsdurthreeDB
	\setlength{\fsdurthreeDB}{-3.2mm}
		\begin{adjustbox}{valign=t}
		\tiny
			\begin{tabular}{ccccccccc}
				\includegraphics[width=\widthscaleeight \textwidth]{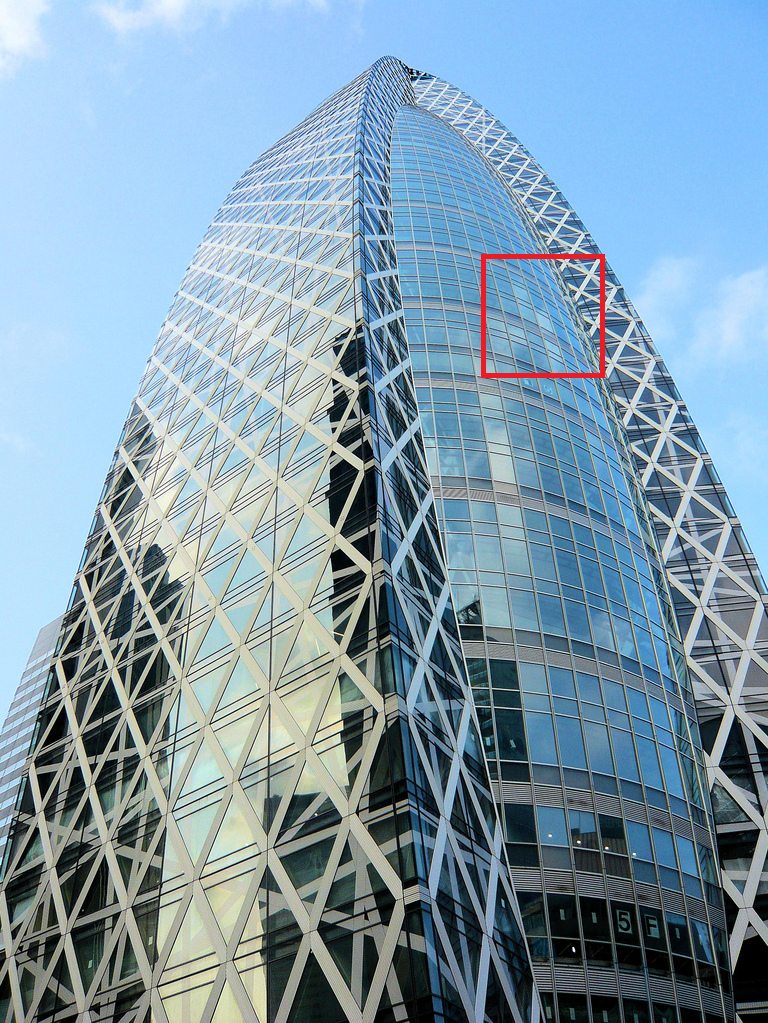} \hspace{\fsdurthreeDB} &
                \includegraphics[width=\widthscaleeight \textwidth]{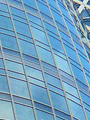} \hspace{\fsdurthreeDB} &
				\includegraphics[width=\widthscaleeight \textwidth]{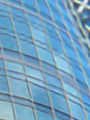} \hspace{\fsdurthreeDB} &
				\includegraphics[width=\widthscaleeight \textwidth]{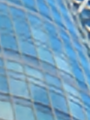} \hspace{\fsdurthreeDB} &
				\includegraphics[width=\widthscaleeight \textwidth]{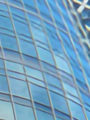} \hspace{\fsdurthreeDB} &

				\includegraphics[width=\widthscaleeight \textwidth]{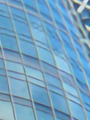} \hspace{\fsdurthreeDB} &
				\includegraphics[width=\widthscaleeight \textwidth]{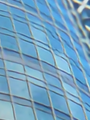} \hspace{\fsdurthreeDB} &
				\includegraphics[width=\widthscaleeight \textwidth]{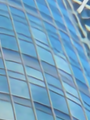} \hspace{\fsdurthreeDB} &
				
				\\
                \emph{Img\_039}~ from Urban100&
				HR~ (PSNR) \hspace{\fsdurthreeDB} &
				EDSR~\cite{lim:CVPR:2017:EDSR}~ (23.97dB) \hspace{\fsdurthreeDB} &
				SRMDNF~\cite{zhang:CVPR:2018SRMDNF}~ (22.96dB) \hspace{\fsdurthreeDB} &
				RDN~\cite{zhang:CVPR:2018RDN}~ (25.32dB) \hspace{\fsdurthreeDB} &
                RCAN~\cite{zhang:ECCV:2018RCAN}~ (24.75dB)\hspace{\fsdurthreeDB} &
				SRFBN~\cite{li:CVPR:2019feedback}~ (25.03dB)\hspace{\fsdurthreeDB} &
				\textbf{Ours~} (\textbf{25.54dB})\hspace{\fsdurthreeDB} &
		
				\\
			\end{tabular}
		\end{adjustbox}
		\vspace{0.5mm}
		
	\caption{
		SISR visual quality comparisons of different methods on `Img\_039' from Urban100 \cite{huang:CVPR:2015Urban100}. The degradation involves Gaussian kernel with kernel width 1.3 and direct downsamping with scale factor $\times3$.
	}
	\label{fig:DB_urban39}
\vspace{-5mm}
\end{figure*}

\begin{figure*}[htbp]
	\scriptsize
	\centering
    \newlength\fsdurthreeDBB
	\setlength{\fsdurthreeDBB}{-3.2mm}
		\begin{adjustbox}{valign=t}
		\tiny
			\begin{tabular}{ccccccccc}
				\includegraphics[angle=90, width=\widthscaleeight \textwidth]{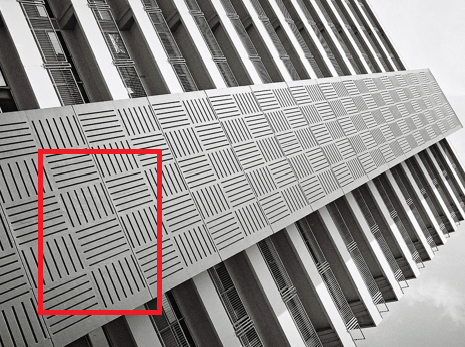} \hspace{\fsdurthreeDBB} &
                \includegraphics[width=\widthscaleeight \textwidth]{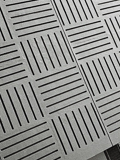} \hspace{\fsdurthreeDBB} &
				\includegraphics[width=\widthscaleeight \textwidth]{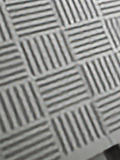} \hspace{\fsdurthreeDBB} &
				\includegraphics[width=\widthscaleeight \textwidth]{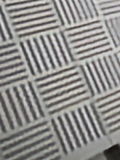} \hspace{\fsdurthreeDBB} &
				\includegraphics[width=\widthscaleeight \textwidth]{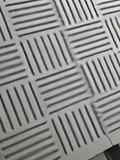} \hspace{\fsdurthreeDBB} &

				\includegraphics[width=\widthscaleeight \textwidth]{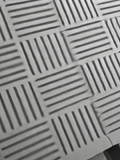} \hspace{\fsdurthreeDBB} &
				\includegraphics[width=\widthscaleeight \textwidth]{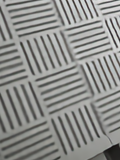} \hspace{\fsdurthreeDBB} &
				\includegraphics[width=\widthscaleeight \textwidth]{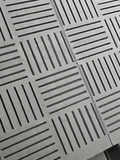} \hspace{\fsdurthreeDBB} &
				
				\\
                \emph{Img\_092}~ from Urban100&
				HR~ (PSNR) \hspace{\fsdurthreeDBB} &
				EDSR~\cite{lim:CVPR:2017:EDSR}~ (20.85dB) \hspace{\fsdurthreeDBB} &
				SRMDNF~\cite{zhang:CVPR:2018SRMDNF}~ (20.51dB) \hspace{\fsdurthreeDBB} &
				RDN~\cite{zhang:CVPR:2018RDN}~ (21.90dB) \hspace{\fsdurthreeDBB} &
                RCAN~\cite{zhang:ECCV:2018RCAN}~ (21.82dB)\hspace{\fsdurthreeDBB} &
				SRFBN~\cite{li:CVPR:2019feedback}~ (21.61dB)\hspace{\fsdurthreeDBB} &
				\textbf{Ours~} (\textbf{25.62dB})\hspace{\fsdurthreeDBB} &
		
				\\
			\end{tabular}
		\end{adjustbox}
		\vspace{0.5mm}
		
	\caption{
		SISR visual quality comparisons of different methods on `Img\_092' from Urban100 \cite{huang:CVPR:2015Urban100}. The degradation involves Gaussian kernel with kernel width 2.6 and direct downsamping with scale factor $\times2$.
	}
	\label{fig:DB_urban92}
\end{figure*}

Tab. \ref{tab:deblur} shows the average PSNR results of three different Gaussian kernel widths $0.5, 1.3, 2.6$ and three different scaling factors $\times2, \times3, \times4$ respectively. Four commonly used benchmark datasets are adopted to verify the effectiveness of  our method. Tab.~\ref{tab:deblur} demonstrates that our method can achieve better performance than other competing methods. One can be seen that our method handles multiple degradations better than others. It is worth mentioning that the performance for the kernel width of $0.5$ is not as good as that for the kernel width of $1.3$. One possible explanation is that more information might get lost by down-sampling when Gaussian kernel width is $0.5$.
The visual quality comparisons are given in Fig.~\ref{fig:DB_urban39} and Fig. \ref{fig:DB_urban92}.
Fig. \ref{fig:DB_urban39} shows that SRFBN \cite{li:CVPR:2019feedback} and our method can recover sharper  edges than other methods. However, SRFBN \cite{li:CVPR:2019feedback} generates more twisty artifacts than ours.
From Fig. \ref{fig:DB_urban92}, we can observe that our method can achieve a clearer image with sharper edges than other benchmark methods, which justifies the superiority of our approach.
Tab. \ref{tab:directly} and Tab. \ref{tab:deblur} show the results of directly downsampling degradation and multiple degradation which degradation matrix lacks structural constraint \cite{dong2013sparse} causing coherence. The reference model, nonlocal autoregressive model(NARM) with improved incoherence properties by connecting a pixel with its nonlocal neighbors has been integrated into our model as trainable \emph{Nonlocal-AR module}. Obviously, our proposed model can achieve better performance than other methods with the benefits of relieving coherence by \emph{Nonlocal-AR module} aiming to break coherence as well as effective model-guild designed network.

\begin{table*}[htbh]

\begin{center}
\centering{
\caption{Average PSNR results for \textbf{multiple} degradation on four benchmark datasets.
 The best performance is shown in bold.}
\label{tab:deblur}
\resizebox{1.8\columnwidth}{!}{
\begin{tabular}{|l|c|c|c|c|c|c|c|c|c|c|c|c|c|}
\hline
\multirow{2}{*}{Method} & Kernel  & \multicolumn{3}{c|}{Set5 \cite{bevilacqua:BMCV:2012Set5}} & \multicolumn{3}{c|}{Set14 \cite{zeyde:CCS:2010Set14}} & \multicolumn{3}{c|}{BSD100 \cite{martin:ICCV:2001BSD100}} & \multicolumn{3}{c|}{Urban100 \cite{huang:CVPR:2015Urban100}} \\  \cline{3-14}
                        & Width                              &$\times$2&$\times$3&$\times$4&$\times$2&$\times$3&$\times$4&$\times$2&$\times$3&$\times$4&$\times$2&$\times$3&$\times$4       \\ \hline \hline
  EDSR\cite{lim:CVPR:2017:EDSR}                      & 0.5       &35.52    &32.37    &29.67    &32.09    &28.63    &26.61   &30.94    &27.69    &26.04    &29.77   &26.03     &23.93     \\
  SRMDNF\cite{zhang:CVPR:2018SRMDNF}                 & 0.5       &35.01    &31.30    &28.83    &31.11    &27.98    &26.04   &30.12    &27.14    &25.55    &27.66   &24.76     &22.93     \\
  RDN\cite{zhang:CVPR:2018RDN}                    & 0.5        &36.46    &32.72    &30.00    &32.07    &28.87    &26.94   &30.95    &27.82    &26.31   &29.89   &26.90     &24.49     \\
  RCAN\cite{zhang:ECCV:2018RCAN}                    & 0.5        &36.37    &32.62    &29.99    &31.94    &28.78    &26.79   &30.87    &27.73    &26.22    &29.49   &26.46     &24.40     \\
  SRFBN\cite{li:CVPR:2019feedback}                   & 0.5       &36.31    &32.56    &29.92    &31.86    &28.72    &26.88   &30.78    &27.75    &26.16    &28.57   &25.67     &24.21     \\
  \textbf{MoG-DUN(ours)}                                                & 0.5       &\textbf{36.60}    &\textbf{32.94}    &\textbf{30.13}    &\textbf{32.42}    &\textbf{29.01}    &\textbf{26.97}   &\textbf{31.09}    &\textbf{28.01}    &\textbf{26.41}    &\textbf{30.53}    &\textbf{27.04}   &\textbf{24.75}          \\

                      \hline \hline
  EDSR\cite{lim:CVPR:2017:EDSR}                      & 1.3       &37.51    &33.87    &31.29    &33.26    &29.97    &28.01   &31.98    &28.84    &27.12    &30.69   &27.45     &25.30     \\
  SRMDNF\cite{zhang:CVPR:2018SRMDNF}                 & 1.3       &35.67    &32.58    &30.17    &31.89    &29.24    &27.31   &31.00    &28.32    &26.71    &28.60   &26.20     &24.30     \\
  RDN\cite{zhang:CVPR:2018RDN}                   & 1.3         &37.45    &34.26    &31.65    &32.31    &30.27    &28.36   &32.05    &29.05    &27.35    &30.94   &28.24     &25.97     \\
  RCAN\cite{zhang:ECCV:2018RCAN}                    & 1.3        &37.44    &34.06    &31.59    &32.18    &30.12    &28.27   &31.96    &28.92    &27.31    &30.57   &27.64     &25.98     \\
  SRFBN\cite{li:CVPR:2019feedback}                   & 1.3       &37.39    &34.09    &31.55    &33.21    &30.08    &28.26   &31.89    &28.87    &27.26    &30.66   &27.84     &25.71     \\
  \textbf{MoG-DUN(ours)}                                                & 1.3       &\textbf{38.11}    &\textbf{34.51}    &\textbf{32.00}    &\textbf{33.87}    &\textbf{30.40}    &\textbf{28.46}   &\textbf{32.19}    &\textbf{29.15}    &\textbf{27.38}    &\textbf{31.73}    &\textbf{28.42}   &\textbf{26.22}     \\
                       \hline \hline

  EDSR\cite{lim:CVPR:2017:EDSR}                      &2.6        &33.79    &33.07    &31.47    &30.09    &29.46    &28.18   &29.02    &28.48    &27.27    &26.89   &26.69     &25.22     \\
  SRMDNF\cite{zhang:CVPR:2018SRMDNF}                 & 2.6       &32.28    &31.83    &30.32    &29.09    &28.72    &27.45   &28.34    &27.94    &26.87    &25.62   &25.69    &24.44     \\
  RDN\cite{zhang:CVPR:2018RDN}                   & 2.6         &33.61    &33.50    &32.04    &30.02    &29.73    &28.56   &29.04    &28.65    &27.57    &26.93   &27.23     &25.82     \\
  RCAN\cite{zhang:ECCV:2018RCAN}                   & 2.6        &33.70    &33.28    &31.99    &29.91    &29.55    &28.50   &28.98    &28.51    &27.53    &26.80   &26.85     &25.79     \\
  SRFBN\cite{li:CVPR:2019feedback}                   & 2.6       &33.64    &33.15    &31.74    &29.94    &29.47    &28.36   &28.89    &28.37    &27.38    &26.78   &26.81     &25.66     \\
  \textbf{MoG-DUN(ours)}                                                & 2.6       &\textbf{37.91}    &\textbf{34.64}    &\textbf{32.48}    &\textbf{33.61}    &\textbf{30.52}    &\textbf{28.85}   &\textbf{32.30}    &\textbf{29.26}    &\textbf{27.71}    &\textbf{30.85}   &\textbf{28.06}    &\textbf{26.22}    \\
                         \hline
\end{tabular}
}}
\vspace{-4mm}
\end{center}
\end{table*}

\vspace{-0.2in}


\subsection{Cost-Performance Tradeoff}
To demonstrate the trade-off between the cost (in terms of the number of parameters) and the performance (as measured by PSNR values), we have compared this work against nine existing SISR methods in Fig.\ref{fig:parameters}. We can see that our model can achieve better PSNR performance than those models with comparable model size (e.g., SRFBN \cite{li:CVPR:2019feedback} and DPDNN \cite{dong:TPAMI:2018DPDNN}) or significant cost savings over those models with comparable PSNR performance (e.g., RCAN \cite{zhang:ECCV:2018RCAN} and RDN \cite{zhang:CVPR:2018RDN}).  In addition, we have compared
the actual running time as well as flops against other competing methods  in Table. \ref{tab:flops and running time}.
The actual running time is the total time of whole Urban100 dataset during the testing and the flops is evaluated on the LR input size $1\times3\times64\times64$. It can be observed that our model has a similar flops-running time performance to RDN \cite{zhang:CVPR:2018RDN}. Although our proposed model does not achieve the best performance in terms of flops and running time, it still have notable advantages over four other competing methods including EDSR\cite{lim:CVPR:2017:EDSR}, DPDNN\cite{dong:TPAMI:2018DPDNN}, SRFBN\cite{li:CVPR:2019feedback}, USRNet\cite{zhang:CVPR:2020deepUSRNet}.

\begin{table*}[htbh]
\begin{center}
\centering{
\caption{The flops and test running time results of $\times2$ bicubic downsampling. }
\label{tab:flops and running time}
\resizebox{1.8\columnwidth}{!}{
\begin{tabular}{|l|c|c|c|c|c|c|c|c|c|}
\hline
Methods      &EDSR\cite{lim:CVPR:2017:EDSR}      &DPDNN\cite{dong:TPAMI:2018DPDNN}      &RDN\cite{zhang:CVPR:2018RDN}    &RCAN\cite{zhang:ECCV:2018RCAN}   &D-DPBN\cite{haris2018deep}   &SRFBN\cite{li:CVPR:2019feedback}  &USRNet\cite{zhang:CVPR:2020deepUSRNet} &\textbf{MoG-DUN(ours)}   \\  \hline
flops(G)     &166.9     &113.1      &90.6   &62.9   &61.8     &89.7  &151.9  &96.4    \\ \hline
testing time(s)    &27.25      &28.70    &20.76   &15.71   &17.03    &31.15  &82.59 &21.46  \\ \hline

\end{tabular}
}}
\end{center}
\end{table*}

\section{Conclusion}
\label{sec:5}

In this paper, we have demonstrated how to unfold the existing NARM into a multi-stage network implementation that is both explainable and efficient. The unfolded network consists of a concatenation of multi-stage building blocks each of which is decomposed of a deep denoising module, a fast nonlocal-AR module, and a versatile reconstruction module. This work extends the previous work DPDNN \cite{dong:TPAMI:2018DPDNN} in the following aspects. First, the new regularization term characterized by NARM  leads to a three-way (instead of double-headed) alternating optimization, which in principle is applicable to other forms of regularization functions. Meantime, the improved incoherence property of NARM makes it suitable for SISR applications particularly on suppressing aliasing artifacts. Second, the unfolded network allows the hidden states of previous stages to be exploited by the later stage. Such densely connected recurrent network architecture is shown important to the recovery of missing high-frequency information in SISR. Extensive experimental results have been reported to show that our MoG-DUN is capable of achieving an improved trade-off between the cost (in terms of network parameter size) and the performance (in terms of both subjective and objective qualities of reconstructed SR images). Currently, we are exploring further improvement based on recently developed SISR method focusing on learning high-level features via densely residue Laplacian network \cite{anwar2020densely}.

\bibliographystyle{IEEEtran}
\bibliography{ning}

\begin{thebibliography}{10}
\providecommand{\url}[1]{#1}
\csname url@samestyle\endcsname
\providecommand{\newblock}{\relax}
\providecommand{\bibinfo}[2]{#2}
\providecommand{\BIBentrySTDinterwordspacing}{\spaceskip=0pt\relax}
\providecommand{\BIBentryALTinterwordstretchfactor}{4}
\providecommand{\BIBentryALTinterwordspacing}{\spaceskip=\fontdimen2\font plus
\BIBentryALTinterwordstretchfactor\fontdimen3\font minus
  \fontdimen4\font\relax}
\providecommand{\BIBforeignlanguage}[2]{{%
\expandafter\ifx\csname l@#1\endcsname\relax
\typeout{** WARNING: IEEEtran.bst: No hyphenation pattern has been}%
\typeout{** loaded for the language `#1'. Using the pattern for}%
\typeout{** the default language instead.}%
\else
\language=\csname l@#1\endcsname
\fi
#2}}
\providecommand{\BIBdecl}{\relax}
\BIBdecl

\bibitem{dong:ECCV:2014learning}
C.~{Dong}, C.~C. {Loy}, K.~{He}, and X.~{Tang}, ``Learning a deep convolutional
  network for image super-resolution,'' in \emph{European Conference on
  Computer Vision}, 2014, pp. 184--199.

\bibitem{ledig2017photo}
C.~Ledig, L.~Theis, F.~Husz{\'a}r, J.~Caballero, A.~Cunningham, A.~Acosta,
  A.~Aitken, A.~Tejani, J.~Totz, Z.~Wang \emph{et~al.}, ``Photo-realistic
  single image super-resolution using a generative adversarial network,'' in
  \emph{Proceedings of the IEEE conference on computer vision and pattern
  recognition}, 2017, pp. 4681--4690.

\bibitem{kim2016accurate}
J.~Kim, J.~Kwon~Lee, and K.~Mu~Lee, ``Accurate image super-resolution using
  very deep convolutional networks,'' in \emph{Proceedings of the IEEE
  conference on computer vision and pattern recognition}, 2016, pp. 1646--1654.

\bibitem{chen:TPAMI:2017TNRD}
Y.~Chen and T.~Pock, ``Trainable nonlinear reaction diffusion: A flexible
  framework for fast and effective image restoration,'' \emph{IEEE Transactions
  on Pattern Analysis Machine Intelligence}, vol.~39, no.~6, pp. 1256--1272,
  2017.

\bibitem{kim:CVPR:2016DRCN}
J.~{Kim}, J.~K. {Lee}, and K.~M. {Lee}, ``Deeply-recursive convolutional
  network for image super-resolution,'' in \emph{2016 IEEE Conference on
  Computer Vision and Pattern Recognition (CVPR)}, 2016, pp. 1637--1645.

\bibitem{lim:CVPR:2017:EDSR}
B.~{Lim}, S.~{Son}, H.~{Kim}, S.~{Nah}, and K.~M. {Lee}, ``Enhanced deep
  residual networks for single image super-resolution,'' in \emph{2017 IEEE
  Conference on Computer Vision and Pattern Recognition Workshops (CVPRW)},
  2017, pp. 1132--1140.

\bibitem{lai2017deep}
W.-S. Lai, J.-B. Huang, N.~Ahuja, and M.-H. Yang, ``Deep laplacian pyramid
  networks for fast and accurate super-resolution,'' in \emph{Proceedings of
  the IEEE conference on computer vision and pattern recognition}, 2017, pp.
  624--632.

\bibitem{haris2018deep}
M.~Haris, G.~Shakhnarovich, and N.~Ukita, ``Deep back-projection networks for
  super-resolution,'' in \emph{Proceedings of the IEEE conference on computer
  vision and pattern recognition}, 2018, pp. 1664--1673.

\bibitem{zhang:CVPR:2018RDN}
Y.~Zhang, Y.~Tian, Y.~Kong, B.~Zhong, and Y.~Fu, ``Residual dense network for
  image super-resolution,'' in \emph{Proceedings of the IEEE Conference on
  Computer Vision and Pattern Recognition}, 2018, pp. 2472--2481.

\bibitem{zhang:ECCV:2018RCAN}
Y.~Zhang, K.~Li, K.~Li, L.~Wang, B.~Zhong, and Y.~Fu, ``Image super-resolution
  using very deep residual channel attention networks,'' in \emph{Proceedings
  of the European Conference on Computer Vision (ECCV)}, 2018, pp. 286--301.

\bibitem{dai2019second}
T.~Dai, J.~Cai, Y.~Zhang, S.-T. Xia, and L.~Zhang, ``Second-order attention
  network for single image super-resolution,'' in \emph{Proceedings of the IEEE
  conference on computer vision and pattern recognition}, 2019, pp.
  11\,065--11\,074.

\bibitem{zhang:CVPR:2018SRMDNF}
K.~{Zhang}, W.~{Zuo}, and L.~{Zhang}, ``Learning a single convolutional
  super-resolution network for multiple degradations,'' in \emph{2018 IEEE/CVF
  Conference on Computer Vision and Pattern Recognition}, 2018, pp. 3262--3271.

\bibitem{li:CVPR:2019feedback}
Z.~{Li}, J.~{Yang}, Z.~{Liu}, X.~{Yang}, G.~{Jeon}, and W.~{Wu}, ``Feedback
  network for image super-resolution,'' in \emph{Proceedings of the IEEE
  Conference on Computer Vision and Pattern Recognition}, 2019, pp. 3867--3876.

\bibitem{liu2020residual}
J.~Liu, W.~Zhang, Y.~Tang, J.~Tang, and G.~Wu, ``Residual feature aggregation
  network for image super-resolution,'' in \emph{Proceedings of the IEEE/CVF
  Conference on Computer Vision and Pattern Recognition}, 2020, pp. 2359--2368.

\bibitem{molnar2020interpretable}
C.~Molnar, \emph{Interpretable machine learning}.\hskip 1em plus 0.5em minus
  0.4em\relax Lulu. com, 2020.

\bibitem{ahn2018fast}
N.~Ahn, B.~Kang, and K.-A. Sohn, ``Fast, accurate, and lightweight
  super-resolution with cascading residual network,'' in \emph{Proceedings of
  the European Conference on Computer Vision (ECCV)}, 2018, pp. 252--268.

\bibitem{wisdom2017building}
S.~Wisdom, T.~Powers, J.~Pitton, and L.~Atlas, ``Building recurrent networks by
  unfolding iterative thresholding for sequential sparse recovery,'' in
  \emph{2017 IEEE International Conference on Acoustics, Speech and Signal
  Processing (ICASSP)}.\hskip 1em plus 0.5em minus 0.4em\relax IEEE, 2017, pp.
  4346--4350.

\bibitem{bertocchi2019deep}
C.~Bertocchi, E.~Chouzenoux, M.-C. Corbineau, J.-C. Pesquet, and M.~Prato,
  ``Deep unfolding of a proximal interior point method for image restoration,''
  \emph{Inverse Problems}, 2019.

\bibitem{hershey2014deep}
J.~R. Hershey, J.~L. Roux, and F.~Weninger, ``Deep unfolding: Model-based
  inspiration of novel deep architectures,'' \emph{arXiv preprint
  arXiv:1409.2574}, 2014.

\bibitem{dong:TPAMI:2018DPDNN}
W.~Dong, P.~Wang, W.~Yin, G.~Shi, F.~Wu, and X.~Lu, ``Denoising prior driven
  deep neural network for image restoration,'' \emph{IEEE transactions on
  pattern analysis and machine intelligence}, vol.~41, no.~10, pp. 2305--2318,
  2018.

\bibitem{zhang:CVPR:2020deepUSRNet}
K.~Zhang, L.~Van~Gool, and R.~Timofte, ``Deep unfolding network for image
  super-resolution,'' in \emph{IEEE Conference on Computer Vision and Pattern
  Recognition}, 2020, pp. 3217--3226.

\bibitem{buades:CVPR:2005nonlocal}
A.~{Buades}, B.~{Coll}, and J.-M. {Morel}, ``A non-local algorithm for image
  denoising,'' in \emph{2005 IEEE Computer Society Conference on Computer
  Vision and Pattern Recognition (CVPR'05)}, vol.~2, no.~2, 2005, pp. 60--65.

\bibitem{yang:TIP:2010:sparse}
J.~{Yang}, J.~{Wright}, T.~S. {Huang}, and Y.~{Ma}, ``Image super-resolution
  via sparse representation,'' \emph{IEEE Transactions on Image Processing},
  vol.~19, no.~11, pp. 2861--2873, 2010.

\bibitem{dong:TIP:2013nonlocally}
W.~{Dong}, L.~{Zhang}, G.~{Shi}, and X.~{Li}, ``Nonlocally centralized sparse
  representation for image restoration,'' \emph{IEEE Transactions on Image
  Processing}, vol.~22, no.~4, pp. 1620--1630, 2013.

\bibitem{kim:TPAMI:2010sparse}
K.~I. {Kim} and Y.~{Kwon}, ``Single-image super-resolution using sparse
  regression and natural image prior,'' \emph{IEEE Transactions on Pattern
  Analysis and Machine Intelligence}, vol.~32, no.~6, pp. 1127--1133, 2010.

\bibitem{dong2013sparse}
W.~Dong, L.~Zhang, R.~Lukac, and G.~Shi, ``Sparse representation based image
  interpolation with nonlocal autoregressive modeling,'' \emph{IEEE
  Transactions on Image Processing}, vol.~22, no.~4, pp. 1382--1394, 2013.

\bibitem{le2011optimization}
Q.~V. Le, J.~Ngiam, A.~Coates, A.~Lahiri, B.~Prochnow, and A.~Y. Ng, ``On
  optimization methods for deep learning,'' in \emph{ICML}, 2011.

\bibitem{adcock2013breaking}
B.~Adcock, A.~C. Hansen, C.~Poon, B.~Roman \emph{et~al.}, ``Breaking the
  coherence barrier: asymptotic incoherence and asymptotic sparsity in
  compressed sensing,'' \emph{arXiv preprint arXiv:1302.0561}, 2013.

\bibitem{he2013half}
R.~He, W.-S. Zheng, T.~Tan, and Z.~Sun, ``Half-quadratic-based iterative
  minimization for robust sparse representation,'' \emph{IEEE transactions on
  pattern analysis and machine intelligence}, vol.~36, no.~2, pp. 261--275,
  2013.

\bibitem{zhang2017beyond}
K.~Zhang, W.~Zuo, Y.~Chen, D.~Meng, and L.~Zhang, ``Beyond a gaussian denoiser:
  Residual learning of deep cnn for image denoising,'' \emph{IEEE Transactions
  on Image Processing}, vol.~26, no.~7, pp. 3142--3155, 2017.

\bibitem{zhang:CVPR:2017learning}
K.~{Zhang}, W.~{Zuo}, S.~{Gu}, and L.~{Zhang}, ``Learning deep cnn denoiser
  prior for image restoration,'' \emph{arXiv preprint arXiv:1704.03264}, 2017.

\bibitem{saab2008stable}
R.~Saab, R.~Chartrand, and O.~Yilmaz, ``Stable sparse approximations via
  nonconvex optimization,'' in \emph{IEEE international conference on
  acoustics, speech and signal processing}, 2008, pp. 3885--3888.

\bibitem{jain2017non}
P.~Jain, P.~Kar \emph{et~al.}, ``Non-convex optimization for machine
  learning,'' \emph{Foundations and Trends{\textregistered} in Machine
  Learning}, vol.~10, no. 3-4, pp. 142--363, 2017.

\bibitem{goodfellow2016deep}
I.~Goodfellow, Y.~Bengio, and A.~Courville, \emph{Deep learning}.\hskip 1em
  plus 0.5em minus 0.4em\relax MIT press, 2016.

\bibitem{boyd2011distributed}
S.~Boyd, N.~Parikh, and E.~Chu, \emph{Distributed optimization and statistical
  learning via the alternating direction method of multipliers}.\hskip 1em plus
  0.5em minus 0.4em\relax Now Publishers Inc, 2011.

\bibitem{he:CVPR:2016deep}
K.~{He}, X.~{Zhang}, S.~{Ren}, and J.~{Sun}, ``Deep residual learning for image
  recognition,'' in \emph{2016 IEEE Conference on Computer Vision and Pattern
  Recognition (CVPR)}, 2016, pp. 770--778.

\bibitem{huang2017densely}
G.~Huang, Z.~Liu, L.~Van Der~Maaten, and K.~Q. Weinberger, ``Densely connected
  convolutional networks,'' in \emph{Proceedings of the IEEE conference on
  computer vision and pattern recognition}, 2017, pp. 4700--4708.

\bibitem{gregor2015draw}
K.~Gregor, I.~Danihelka, A.~Graves, D.~J. Rezende, and D.~Wierstra, ``Draw: A
  recurrent neural network for image generation,'' \emph{arXiv preprint
  arXiv:1502.04623}, 2015.

\bibitem{tai:CVPR:2017DRRN}
Y.~Tai, J.~Yang, and X.~Liu, ``Image super-resolution via deep recursive
  residual network,'' in \emph{Proceedings of the IEEE conference on computer
  vision and pattern recognition}, 2017, pp. 3147--3155.

\bibitem{zhao:IJCAI:2019recurrent}
Y.~Zhao, Y.~Shen, and J.~Yao, ``Recurrent neural network for text
  classification with hierarchical multiscale dense connections.'' in
  \emph{IJCAI}, 2019, pp. 5450--5456.

\bibitem{indyk1998approximate}
P.~Indyk and R.~Motwani, ``Approximate nearest neighbors: towards removing the
  curse of dimensionality,'' in \emph{Proceedings of the thirtieth annual ACM
  symposium on Theory of computing}, 1998, pp. 604--613.

\bibitem{wang:cvpr:2018non-local}
X.~Wang, R.~Girshick, A.~Gupta, and K.~He, ``Non-local neural networks,'' in
  \emph{Proceedings of the IEEE conference on computer vision and pattern
  recognition}, 2018, pp. 7794--7803.

\bibitem{liu:nips:2018non-local}
D.~Liu, B.~Wen, Y.~Fan, C.~C. Loy, and T.~S. Huang, ``Non-local recurrent
  network for image restoration,'' in \emph{Advances in Neural Information
  Processing Systems}, 2018, pp. 1673--1682.

\bibitem{tomasi1998bilateral}
C.~Tomasi and R.~Manduchi, ``Bilateral filtering for gray and color images,''
  in \emph{Sixth international conference on computer vision (IEEE Cat. No.
  98CH36271)}.\hskip 1em plus 0.5em minus 0.4em\relax IEEE, 1998, pp. 839--846.

\bibitem{yue2018compact}
K.~Yue, M.~Sun, Y.~Yuan, F.~Zhou, E.~Ding, and F.~Xu, ``Compact generalized
  non-local network,'' in \emph{Advances in Neural Information Processing
  Systems}, 2018, pp. 6510--6519.

\bibitem{timofte:CVPRW:2017DIV2K}
R.~Timofte, E.~Agustsson, L.~Van~Gool, M.-H. Yang, and L.~Zhang, ``Ntire 2017
  challenge on single image super-resolution: Methods and results,'' in
  \emph{Proceedings of the IEEE Conference on Computer Vision and Pattern
  Recognition Workshops}, 2017, pp. 114--125.

\bibitem{bevilacqua:BMCV:2012Set5}
M.~{Bevilacqua}, A.~{Roumy}, C.~{Guillemot}, and M.~L. {Alberi-Morel},
  ``Low-complexity single-image super-resolution based on nonnegative neighbor
  embedding,'' in \emph{British Machine Vision Conference 2012}, 2012, pp.
  1--10.

\bibitem{zeyde:CCS:2010Set14}
R.~{Zeyde}, M.~{Elad}, and M.~{Protter}, ``On single image scale-up using
  sparse-representations,'' in \emph{Proceedings of the 7th international
  conference on Curves and Surfaces}, 2010, pp. 711--730.

\bibitem{martin:ICCV:2001BSD100}
D.~{Martin}, C.~{Fowlkes}, D.~{Tal}, and J.~{Malik}, ``A database of human
  segmented natural images and its application to evaluating segmentation
  algorithms and measuring ecological statistics,'' in \emph{Proceedings Eighth
  IEEE International Conference on Computer Vision. ICCV 2001}, vol.~2, 2001,
  pp. 416--423.

\bibitem{huang:CVPR:2015Urban100}
J.-B. {Huang}, A.~{Singh}, and N.~{Ahuja}, ``Single image super-resolution from
  transformed self-exemplars,'' in \emph{2015 IEEE Conference on Computer
  Vision and Pattern Recognition (CVPR)}, 2015, pp. 5197--5206.

\bibitem{matsui:MTA:2017manga}
Y.~{Matsui}, K.~{Ito}, Y.~{Aramaki}, A.~{Fujimoto}, T.~{Ogawa}, T.~{Yamasaki},
  and K.~{Aizawa}, ``Sketch-based manga retrieval using manga109 dataset,''
  \emph{Multimedia Tools and Applications}, vol.~76, no.~20, pp.
  21\,811--21\,838, 2017.

\bibitem{wang:TIP:2004SSIM}
Z.~{Wang}, A.~{Bovik}, H.~{Sheikh}, and E.~{Simoncelli}, ``Image quality
  assessment: from error visibility to structural similarity,'' \emph{IEEE
  Transactions on Image Processing}, vol.~13, no.~4, pp. 600--612, 2004.

\bibitem{kingma2014adam}
D.~P. Kingma and J.~Ba, ``Adam: A method for stochastic optimization,''
  \emph{arXiv preprint arXiv:1412.6980}, 2014.

\bibitem{han:CVPR:2018DSRN}
W.~{Han}, S.~{Chang}, D.~{Liu}, M.~{Yu}, M.~{Witbrock}, and T.~S. {Huang},
  ``Image super-resolution via dual-state recurrent networks,'' in \emph{2018
  IEEE/CVF Conference on Computer Vision and Pattern Recognition}, 2018, pp.
  1654--1663.

\bibitem{xie2020weight}
L.~Xie, X.~Chen, K.~Bi, L.~Wei, Y.~Xu, Z.~Chen, L.~Wang, A.~Xiao, J.~Chang,
  X.~Zhang \emph{et~al.}, ``Weight-sharing neural architecture search: A battle
  to shrink the optimization gap,'' \emph{arXiv preprint arXiv:2008.01475},
  2020.

\bibitem{anwar2020densely}
S.~Anwar and N.~Barnes, ``Densely residual laplacian super-resolution,''
  \emph{IEEE Transactions on Pattern Analysis and Machine Intelligence}, 2020.

\end{thebibliography}

\end{document}